\documentclass[letterpaper]{article} 
\usepackage{aaai2026}  
\usepackage{times}  
\usepackage{helvet}  
\usepackage{courier}  
\usepackage[hyphens]{url}  
\usepackage{graphicx} 
\usepackage{natbib}  
\usepackage{caption} 
\usepackage{algorithm}
\usepackage{algorithmic}

\usepackage{newfloat}
\usepackage{listings}
\DeclareCaptionStyle{ruled}{labelfont=normalfont,labelsep=colon,strut=off} 
\floatstyle{ruled}
\newfloat{listing}{tb}{lst}{}
\floatname{listing}{Listing}
\title{Deblur4DGS: 4D Gaussian Splatting from Blurry Monocular Videos}
\author{
    Renlong Wu, Zhilu Zhang\thanks{Corresponding author.}, Mingyang Chen,  Zifei Yan, Wangmeng Zuo \\
}
\affiliations{
    Harbin Institute of Technology\\
    { hirenlongwu@gmail.com,cszlzhang@outlook.com, youngmchan269@gmail.com,} \\
    { yanzifei@hit.edu.cn,wmzuo@hit.edu.cn}\\
}

\usepackage{bibentry}

\makeatletter
\DeclareRobustCommand\onedot{\futurelet\@let@token\@onedot}
\def\@onedot{\ifx\@let@token.\else.\null\fi\xspace}
\def\eg{\emph{e.g}\onedot} 
\def\ie{\emph{i.e}\onedot}

\makeatother

\newcommand{\tabincell}[2]{\begin{tabular}{@{}#1@{}}#2\end{tabular}}
\usepackage{xcolor}

\usepackage{booktabs, multirow, multicol, makecell}
\usepackage{graphicx}
\usepackage{amssymb}
\usepackage{xcolor}
\usepackage[abs]{overpic}
\usepackage{amsmath}
\RequirePackage{xspace}
\usepackage{cleveref} 

\usepackage{subcaption} 
\usepackage{array}      

\begin{document}

\maketitle

\begin{abstract}
Recent 4D reconstruction methods have yielded impressive results but rely on sharp videos as supervision. However, motion blur often occurs in videos due to camera shake and object movement, while existing methods render blurry results when using such videos for reconstructing 4D models. Although a few approaches attempted to address the problem, they struggled to produce high-quality results, due to the inaccuracy in estimating continuous dynamic representations within the exposure time. Encouraged by recent works in 3D motion trajectory modeling using 3D Gaussian Splatting (3DGS), we take 3DGS as the scene representation manner, and propose Deblur4DGS to reconstruct a high-quality 4D model from blurry monocular video. Specifically, we transform continuous dynamic representations estimation within an exposure time into the exposure time estimation. Moreover, we introduce the exposure
regularization term, multi-frame, and multi-resolution consistency regularization term to avoid trivial solutions. Furthermore, to better represent objects with large motion, we suggest blur-aware variable canonical Gaussians. Beyond novel-view synthesis, Deblur4DGS can be applied to improve blurry video from multiple perspectives, including deblurring, frame interpolation, and video stabilization. Extensive experiments in both synthetic and real-world data on the above four tasks show that Deblur4DGS outperforms state-of-the-art 4D reconstruction methods. The codes are available at \url{https://github.com/ZcsrenlongZ/Deblur4DGS}.
\end{abstract}


\section{Introduction}
Substantial efforts have been made for 4D reconstruction, which has extensive applications in augmented reality and virtual reality.  
To model static scenes, Neural Radiance Field (NeRF)~\cite{mildenhall2021nerf} and 3D Gaussian Splatting (3DGS)~\cite{kerbl20233d} propose implicit neural representation manner and explicit Gaussian ellipsoids one, respectively.
To model dynamic objects, implicit neural fields~\cite{zhu2024motiongs,yang2024deformable,4dgaussians,yan2023nerf} and explicit deformation~\cite{4d-rotor20244d,chu2024dreamscene4d,katsumata2024compact,lin2024gaussian,li2024spacetime,wang2024shape} are suggested for motion representation.
While achieving great progress, most methods rely on synchronized multi-view videos.
They yield unsatisfactory results when applied to monocular video, where dynamic objects are only observed once at each timestamp. 
To alleviate the under-constrained nature of the problem, recent studies have introduced data-driven priors, such as depth maps~\cite{lee2023fast,yang48953364d}, optical flows~\cite{,gao2024gaussianflow,zhu2024motiongs}, tracks~\cite{seidenschwarz2024dynomo,lei2024mosca}, and generative models~\cite{wu2025sc4d, chu2024dreamscene4d,zeng2025stag4d} for better 4D reconstruction.

Unfortunately, motion blur often arises due to camera shake and object movement.
When reconstructing the 4D scene from the blurry video, the above methods usually render blurry results.
The first step to solving this problem is to deal with camera motion blur, which is relatively simple. 
Some NeRF-based~\cite{lee2023dp,wang2023bad,lee2023exblurf} and 3DGS-based~\cite{zhao2024bad,chen2024deblur,oh2024deblurgs} methods have suggested jointly optimizing 3D representation and camera poses within the exposure time by calculating the reconstruction loss between the synthetic blurry images and the input blurry frames.
In contrast, the object motion blur is more challenging to address, as the solution has to estimate continuous and sharp dynamic representations within the exposure time to simulate blurry frames.

In this work, we take 3DGS~\cite{kerbl20233d} as the scene representation manner to explore the problem, driven by two main motivations.
First, its successful application in 4D reconstruction make this method highly promising.
Second, the explicit 3D motion modeling presents an opportunity to simplify the complex continuous dynamic representations estimation within the exposure time into exposure time estimation, avoiding complex extra motion modeling in DyBluRF~\cite{sun2024dyblurf, bui2023dyblurf}.
Once the exposure time is estimated, continuous dynamic representations can be obtained by directly interpolating between representations at the nearest integer timestamps.
We note that the concurrent work BARD-GS~\cite{lu2025bard} adopts a similar strategy, but they perform unsatisfactorily due to the under-constrained optimization, especially for large object motion.

Specifically, we propose Deblur4DGS, a Gaussian Splatting framework for 4D reconstruction from blurry monocular video.
For the static scene, we jointly optimize the  camera poses at exposure start and end with static Gaussians.
For the dynamic objects, we optimize learnable exposure time parameters and dynamic Gaussians of the integer timestamps, simultaneously.
Then continuous camera poses and dynamic Gaussians within exposure time can be obtained by interpolation, and they are used to render continuous sharp frames to calculate the reconstruction loss.
Moreover, to avoid trivial solutions, we introduce the exposure regularization term, as well as the multi-frame and multi-resolution consistency regularization terms.
Furthermore, existing 4D reconstruction methods generally select Gaussians at a single timestamp as canonical Gaussians. 
However, it may produce results with missing details in scenes with large motion, especially when processing blurry videos with a low frame rate.
To alleviate this issue, we suggest variable canonical Gaussians as time progresses based on the image blur level.
Gaussians corresponding to the sharper frame are selected as the canonical ones for better blur removal, and each canonical Gaussian is only used for nearby timestamps to reduce difficulty of modeling large motion.

Blurry videos suffer from not only motion blur, but also low frame rates and scene shake generally.
Beyond novel-view synthesis, the optimized Deblur4DGS can be applied to address these problems, achieving deblurring, frame interpolation, and video stabilization.
We evaluate Deblur4DGS from all four perspectives.
Extensive experiments on both synthetic and real-world data demonstrate that Deblur4DGS outperforms state-of-the-art 4D reconstruction methods quantitatively and qualitatively while maintaining real-time rendering speed.
Furthermore, Deblur4DGS has competitive capabilities in comparison with task-specific video processing models trained in a supervised manner.

The main contributions can be summarized as follows:
\begin{itemize}
    \item We propose Deblur4DGS, a 4D Gaussian Splatting framework specially designed to reconstruct a high-quality 4D model from blurry monocular video. 
    \item We propose transforming dynamic representation estimation into exposure time estimation, where a series of regularizations are suggested to tackle under-constrained optimization and blur-aware variable canonical Gaussians is present to better represent dynamic objects.
    \item Extensive experiments in synthetic and real-world data show that Deblur4DGS outperforms state-of-the-art 4D reconstruction methods on novel-view synthesis, deblurring, frame interpolation, and video stabilization tasks.
    
\end{itemize}

\section{Related Work}

\subsection{Image and Video Deblurring}
Deep learning-based image~\cite{ren2023multiscale,li2023self,wang2022uformer,zhang2022self,zhang2024bracketing} and video~\cite{ zhong2023real,Pan_2023_CVPR, zhong2023blur,zhong2020efficient,chan2022basicvsr++} deblurring methods have been widely explored.
Compared to image deblurring methods, video ones leverage temporal clues between consecutive frames for more effective restoration.
DSTNet~\cite{Pan_2023_CVPR} develops a deep discriminative spatial and temporal network.
BasicVSR++~\cite{chan2022basicvsr++} improves feature fusion with second-order feature propagation and flow-guided alignment.
BSSTNet~\cite{zhang2024blur} introduces a blur map to sufficiently utilize the entire video, achieving recent state-of-the-art.
When reconstructing from a blurry video, pre-processing it with the 2D deblurring method is a straightforward manner.
However, 2D deblurring methods cannot perceive 3D structures and maintain scene geometric consistency, leading to unsatisfactory scene reconstruction.

\begin{figure*}[t!]
    \centering
    \includegraphics[width=0.99\linewidth]{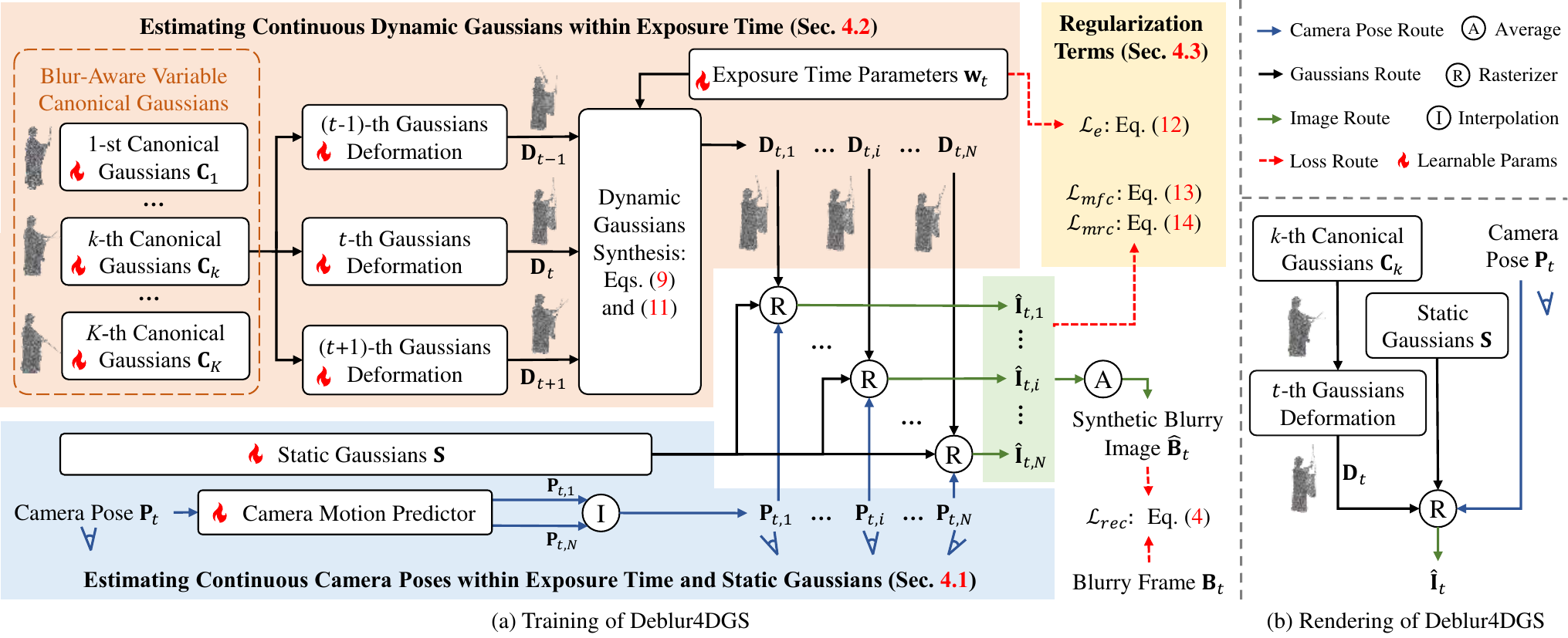}
    \vspace{-3mm}
    \caption{
    (a) Training of Deblur4DGS.
    When processing $t$-th frame, we first discretize its exposure time into $N$ timestamps.
    Then,  we estimate continuous camera poses $\{\mathbf{P}_{t,i}\}_{i=1}^{N}$ and dynamic Gaussians $\{\mathbf{D}_{t,i}\}_{i=1}^{N}$ within exposure time.
    Next, we render each latent sharp image $\hat{\mathbf{I}}_{t,i}$ with the camera pose $\mathbf{P}_{t,i}$, dynamic Gaussians $\mathbf{D}_{t,i}$ and static Gaussians $\mathbf{S}$.
    Finally, $\{\hat{\mathbf{I}}_{t,i}\}_{i=1}^{N}$ are averaged to obtain the synthetic blurry image $\hat{\mathbf{B}}_{t}$, which is used to calculate the reconstruction loss $\mathcal{L}_{rec}$ with the given blurry frame $\mathbf{B}_{t}$.
     To regularize the under-constrained optimization, we introduce  exposure regularization $\mathcal{L}_{e}$, multi-frame consistency regularization $\mathcal{L}_{mfc}$ and multi-resolution consistency regularization $\mathcal{L}_{mrc}$.
    (b) Rendering of Deblur4DGS.  Deblur4DGS produces the sharp image with user-provided timestamp $t$ and camera pose $\mathbf{P}_{t}$.
    }
    \label{fig:pipeline}
\end{figure*}

\subsection{3D and 4D Reconstruction}
To reconstruct 3D models, NeRF~\cite{mildenhall2021nerf} and 3DGS~\cite{kerbl20233d} introduce implicit neural representation manner and explicit Gaussian ellipsoids one respectively, where the latter generally achieves better results.
To reconstruct 4D models, most works~\cite{somraj2024factorized,duan20244d,lu20243d,lin2024gaussian,li2024spacetime,sun20243dgstream,4dgaussians,yang2024deformable,mihajlovic2024splatfields,wang2025gflow} incorporate implicit neural fields and explicit deformation for motion representation.
Moreover, to better reconstruct from monocular video, some studies enhance 4D reconstruction with data-driven priors, such as depth maps~\cite{lee2023fast,yang48953364d}, optical flows~\cite{gao2024gaussianflow,wang2025gflow}, tracks~\cite{wang2024shape,seidenschwarz2024dynomo}, and generative models~\cite{wu2025sc4d, chu2024dreamscene4d}.
For example, GFlow~\cite{wang2025gflow} utilizes only 2D priors to lift a video to a 4D scene.
GaussianMarbles~\cite{stearns2024dynamic} reduces the degrees of freedom of each Gaussian.
Note that these methods heavily rely on high-quality sharp videos for supervision and perform poorly when facing blurry inputs.
To process camera motion in static areas, recent works~\cite{lee2023dp, ma2022deblur,wang2023bad,lee2023exblurf, lee2024deblurring,zhao2024bad,peng2024bags,lee2024crim,chen2024deblur,oh2024deblurgs} suggest jointly optimizing the scene representation and recovering the camera poses within the exposure time.
To process object motion blur in dynamic scenes, DyBluRF~\cite{sun2024dyblurf, bui2023dyblurf}  incorporates object motion blur formation into dynamic model optimization but faces challenges in producing high-quality images and achieving real-time rendering.
In this work, with 3DGS as the scene representation manner, we develop Deblur4DGS to reconstruct a high-quality 4D model from a blurry video.

\section{Preliminary}
\subsection{4D Gaussian Splatting}
\label{section:Preliminary}
A 3D Gaussian~\cite{kerbl20233d} is parameterized by $\{\mathbf{x}, \mathbf{r}, \mathbf{s}, \mathbf{o}, \mathbf{c}\}$, where  $\mathbf{x}$ characterizes the center position in the world space,  rotation matrix $\mathbf{r}$ and scale matrix $\mathbf{s}$ define the shape, $\alpha$ is opacity, and spherical harmonics (SH) coefficients $\mathbf{c}$ represent the view-dependent color. 

4D Gaussian Splatting (4DGS) usually process static and dynamic regions separately.
Static regions can be represented by a set of 3D Gaussians, named $\mathbf{S}$.
For the dynamic areas, 4DGS generally selects a timestamp (\eg, the first timestamp) and represents the objects by canonical dynamic Gaussians, \ie,  $\mathbf{C}$.
Then, $\mathbf{C}$ is deformed to other timestamps for motion representation.
Denote by $\mathbf{D}_{t}$ the dynamic Gaussians at $t$-th timestamp, it can be written as,
\begin{equation}
\mathbf{D}_{t} = \mathcal{F}(\mathbf{C}, t; \Theta_{\mathcal{F}}).
\end{equation}
$\mathcal{F}$ is the deformation operation with parameters $\Theta_{\mathcal{F}}$.
The Gaussians for $t$-th timestamp is the union of  $\mathbf{S}$ and $\mathbf{D}_{t}$.
Collectively, 4DGS models a scene with static Gaussians $\mathbf{S}$, canonical dynamic Gaussians $\mathbf{C}$, and a deformation operation $\mathcal{F}$.
With the provided camera pose, the Gaussians at $t$-th timestamp $\mathbf{D}_{t}$ can be projected into 2D spaces and rasterized to obtain the corresponding image.

\subsection{Motion Blur Formation}
Motion blur occurs due to camera shake and object movement, which can be regarded as the integration of latent sharp images~\cite{nah2017deep}, \ie, 
%
\begin{equation}
\mathbf{B}(u,v)=\phi \int_0^\tau \mathbf{I}_t(u,v) d t.
\end{equation}
$\mathbf{B} \in \mathbb{R}^{H\times W \times 3}$ is the blurry image and $\mathbf{I}_t$ is the latent sharp one at $t$-th timestamp.
$(u,v)$ is pixel location, $\tau$ is the camera exposure time, and $\phi$ is a normalization factor.
To approximate the integral operation, recent works~\cite{zhao2024bad,sun2024dyblurf,wang2023bad} divide the exposure time into $N$ timestamps and regard the blurry image as the average of $N$ sharp images, \ie,
\begin{equation}
\label{eq:formation}
\mathbf{B}(u,v) \approx \frac{1}{N} \sum_{i=0}^{N-1} \mathbf{I}_i(u,v).
\end{equation}

In this work, we reconstruct 4D model from a blurry video by integrating the blur formation into model optimization. 

\section{Proposed Method}
\label{section:overview}
Let $\{\mathbf{B}_{t}\}_{t=1}^{T}$ and $\{\mathbf{M}_{t}\}_{t=1}^{T}$ denote a blurry video with $T$ timestamps and the corresponding masks indicating dynamic areas (extracted by SAM2~\cite{ravi2024sam2}), respectively.
As shown in \cref{fig:pipeline}(a), when processing $t$-th frame, we first evenly divide its camera exposure time into $N$ timestamps.
Then, we estimate continuous camera poses $\{\mathbf{P}_{t,i}\}_{i=1}^{N}$ and dynamic Gaussians $\{\mathbf{D}_{t,i}\}_{i=1}^{N}$ to simulate camera shake and object movement. 
Next, we render each sharp image $\hat{\mathbf{I}}_{t, i}$ with the corresponding camera pose $\mathbf{P}_{t,i}$, dynamic Gaussians $\mathbf{D}_{t,i}$ and static Gaussians $\mathbf{S}$.
After that, we average $\{\hat{\mathbf{I}}_{t,i}\}_{i=1}^{N}$ to obtain the synthetic blurry image $\hat{\mathbf{B}}_{t}$,
which is used to calculate the reconstruction loss $\mathcal{L}_{rec}$ with the given blurry frame $\mathbf{B}_{t}$, \ie,
\begin{equation}
\label{eq:loss_rec}
\mathcal{L}_{rec} = (1-\beta)\mathcal{L}_{1}(\hat{\mathbf{B}}_{t}, \mathbf{B}_{t}) + \beta\mathcal{L}_{ssim}(\hat{\mathbf{B}}_{t}, \mathbf{B}_{t}).
\end{equation}
$\mathcal{L}_{1}$ and $\mathcal{L}_{ssim}$ are $\ell_1 $ loss and SSIM~\cite{wang2004image} loss, respectively.
$\beta$ is set to $0.2$.
The setting all follows 3DGS~\cite{kerbl20233d}.
\subsection{Continuous Camera Poses Estimation}
\label{section:estimate_camera_poses}
To estimate continuous camera poses, recent methods~\cite{zhao2024bad,peng2024bags,chen2024deblur,oh2024deblurgs} directly
optimize exposure start and end poses (\ie, $\mathbf{P}_{t,1}$ and $\mathbf{P}_{t,N}$).
Then, the linear interpolation is performed between $\mathbf{P}_{t,1}$ and $\mathbf{P}_{t,N}$ to obtain the camera pose at $i$-th intermediate timestamp (\ie, $\mathbf{P}_{t,i}$),\ie,
\begin{equation}
   \mathbf{P}_{t,i} = \mathbf{P}_{t,1} \odot \texttt{exp}(\frac{i-1}{N-1} \odot \texttt{log}( \frac{\mathbf{P}_{t, N}}{\mathbf{P}_{t,1}})).
\end{equation}
$\texttt{exp}$ and $\texttt{log}$ are exponential and logarithmic functions, respectively.
$\odot$ is a pixel-wise multiply operation.

We follow the manner but deploy a tiny MLP as the camera motion predictor (see details in the \textit{Suppl}) for more stable optimization.
We pre-train it and static Gaussians $\mathbf{S}$ with static reconstruction loss $\mathcal{L}_{rec}^{s}$, \ie,
\begin{equation}
\label{eq:loss_rec_static}
\mathcal{L}_{rec}^{s} = (1-\beta)\mathcal{L}_{1}(\hat{\mathbf{B}}_{t}^{s}, \mathbf{B}_{t}^{s}) + \beta\mathcal{L}_{ssim}(\hat{\mathbf{B}}_{t}^{s}, \mathbf{B}_{t}^{s}).
\end{equation}
$\hat{\mathbf{B}}_{t}^{s} = (1 - \mathbf{M}_{t}) \odot \hat{\mathbf{B}}_{t}$ and  $\mathbf{B}_{t}^{s} = (1 - \mathbf{M}_{t}) \odot \mathbf{B}_{t}$ are the static areas of $\hat{\mathbf{B}}_{t}$ and $\mathbf{B}_{t}$, respectively.
\subsection{Continuous Dynamic Gaussians Estimation}
\label{sec:continuous_dynamic_objects}  
We first introduce blur-aware variable canonical Gaussians for better dynamic representation at integer timestamps.
Then, we describe Gaussian deformation manner.
Finally, we detail how to take learnable exposure time parameters to obtain continuous dynamic Gaussians within exposure time.
\noindent\textbf{Blur-Aware Variable Canonical Gaussians.} 
Existing 4D reconstruction methods generally select a single canonical Gaussians $\mathbf{C}$ across the entire video, which may produce results with missing details in scenes with large motion.
To alleviate the issue, we suggest varying the canonical Gaussians as time progresses.
In such case, the $k$-th canonical Gaussians $\mathbf{C}_{k}$ is only used for some nearby timestamps, thus reducing the difficulty of motion modeling.
One way to achieve this is to uniformly divide the video into $K$ segments and select $\mathbf{C}_{k}$ for $k$-th segment.
Although it improves performance, selecting the one corresponding to the sharper frame is better for blur removal.
In particular, we first uniformly divide the video into $K$ segments and calculate the blur level  $b_{t}$ of dynamic areas for $t$-th frame following ~\cite{bansal2016blur,ren2020video},
\ie,
%
\begin{equation}
   b_{t} = \sum_{(u,v)\in \mathbf{M}_{t}}(\Delta\mathbf{B}_{t}(u,v) -\overline{\Delta\mathbf{B}_{t}})^2.
\end{equation}
$\mathbf{M}_{t}$ indicates dynamic areas.
$\Delta\mathbf{B}_{t}$ is the image Laplacian and $\overline{\Delta\mathbf{B}_{t}}$ is its mean value.
The larger $b_{t}$ is, the sharper the frame is.
To make the start and end frame of the segment as sharp as possible, we look for the sharp frame among their surrounding $H$ frames and redefine them as the start and end of current segment.
Finally, we select the Gaussians for the sharpest frame in each segment as its canonical ones.

\noindent\textbf{Gaussian  Deformation.}
We deform dynamic Gaussians with a set of rigid transformation matrices, following Shape-of-Motion~\cite{wang2024shape}.
Let $\{\mathbf{x}_{c}, \mathbf{r}_{c}, \mathbf{s}, \mathbf{o}, \mathbf{c}\}$, $\{\mathbf{x}_{t}, \mathbf{r}_{t}, \mathbf{s}, \mathbf{o}, \mathbf{c}\}$, and $\{\mathbf{A}_{t}, \mathbf{E}_{t}\}$ denote a Gaussian in $\mathbf{C}_{k}$, the ones in $\mathbf{D}_{t}$, and the corresponding transformation matrix, respectively.
It can be written as,
\begin{equation}
   \label{eqn:deform}
   \mathbf{x}_{t} = \mathbf{A}_{t} \mathbf{x}_{c} + \mathbf{E}_{t}, \qquad   
   \mathbf{r}_{t} = \mathbf{A}_{t} \mathbf{r}_{c}.
\end{equation}
\noindent\textbf{Interpolation with Exposure Time Parameters.}
To get continuous dynamic Gaussians $\{\mathbf{D}_{t,i}\}_{i=1}^{N}$, one straightforward way is to deploy a series of learnable Gaussian or deformation parameters, but it is unstable to optimize.
With the explicit object motion representation in \cref{eqn:deform}, $\mathbf{D}_{t, i}$ can be calculated by interpolating between the ones at the nearest integer timestamps, \ie,
\begin{equation}
\small
\label{eq:estimate_continuous_gaussians}
\begin{split}
    \mathbf{D}_{t, i} = \mathbf{w}_{t, i} \odot \mathbf{D}_{t-1} + (1 - \mathbf{w}_{t, i}) \odot \mathbf{D}_{t},   \text{ }{i} \in [1, N/2],
    \\
    \mathbf{D}_{t, i} = (1 - \mathbf{w}_{t, i}) \odot \mathbf{D}_{t} + \mathbf{w}_{t, i} \odot \mathbf{D}_{t+1},  \text{ }{i} \in [N/2, N].
\end{split}
\end{equation}
$\mathbf{w}_{t, i}$ is the normalized time interval between $\mathbf{D}_{t, i}$ and $\mathbf{D}_{t, N/2}$.
Thus, the problem is transformed to estimate $\mathbf{w}_{t, i}$.
In the implementation, we can estimate the one at exposure start and end (\ie, $\mathbf{w}_{t, 1}$ and $\mathbf{w}_{t, N}$) and then interpolate between them to get the $i$-th intermediate one $\mathbf{w}_{t, i}$, \ie,
\begin{equation}
    \label{eq:w_t_interpolate}
    \mathbf{w}_{t,i} = (1 - \frac{i-1}{N-1}) \odot \mathbf{w}_{t,1} + \frac{i-1}{N-1} \odot \mathbf{w}_{t,N}.
\end{equation}
As the object motion within the exposure can be regarded as uniform, the absolute value of $\mathbf{w}_{t, 1}$ and $\mathbf{w}_{t, N}$ are equal, which is half the exposure time $\mathbf{w}_{t}$.
Thus, \cref{eq:w_t_interpolate} can be re-written as,
\begin{equation}
\label{eq:w_t_interpolate_final}
    \mathbf{w}_{t,i} = (1 - \frac{i-1}{N-1}) \odot \frac{\mathbf{w}_{t}}{2} + \frac{i-1}{N-1} \odot (-\frac{\mathbf{w}_{t}}{2}).
\end{equation}
Finally, we set learnable parameters $\mathbf{w}_{t}$ for continuous dynamic Gaussians estimation within the exposure time. 
The canonical Gaussians, Gaussian deformation modules, and $\mathbf{w}_{t}$ are jointly optimized.
The reconstruction loss for dynamic areas is similar to \cref{eq:loss_rec_static}.

\subsection{Regularization Terms}
\label{section:regularization_terms}
After optimization with \cref{eq:loss_rec}, static areas of $\hat{\mathbf{I}}_{t, i}$ are sharp while dynamic areas can with notable artifacts.
The reasons are below.
(1) Note that multiple solutions exist for the model to fulfill \cref{eq:loss_rec}.
The most ideal one is that every $\hat{\mathbf{I}}_{t,i}$ is sharp, and the most trivial one is that every $\hat{\mathbf{I}}_{t,i}$ is as blurry as $\mathbf{B}_{t}$.
(2) As static areas are consistent across the entire video, the model tends to learn the underlying sharp representation for inter-frame consistency.
In other words, the inter-frame consistency implicitly regularizes model optimization.
To further validate this, we conduct an experiment that removes the inter-frame consistency by reducing the number of frames to one. 
In such a case, the static areas are blurry after optimization with \cref{eq:loss_rec}, which supports our confirmation.
(3) Compared to static areas, the inter-frame consistency in dynamic ones is weaker due to object motion.
It may provide insufficient regularization to guide sharp representation learning, thus leading to artifacts.
To avoid this, we introduce regularization terms $\mathcal{L}_{reg}$, including exposure regularization $\mathcal{L}_{e}$, multi-frame consistency term $\mathcal{L}_{mfc}$, and multi-resolution consistency term $\mathcal{L}_{mrc}$.
First, the continuous dynamic Gaussians $\{\mathbf{D}_{t, i}\}_{i=1}^{N}$ should not be the same.
In other words, the value of exposure time parameters $\mathbf{w}_{t}$ should not be too small.
If $\mathbf{w}_{t}$ is too small, $\mathbf{D}_{t,i}$ is nearly the same as $\mathbf{D}_{t}$, leading to trivial solutions.
We constrain $\mathbf{w}_{t}$ by $\mathcal{L}_{e}$, as,
\begin{equation}
\label{eq:L_w}
   \mathcal{L}_{e} = \texttt{max}(0,  \mathbf{\epsilon}-\mathbf{w}_{t}).
\end{equation}
$\texttt{max}$ is the maximum function and $\mathbf{\epsilon}$ is a threshold.
Second, despite different motions, the content of multiple frames within exposure time should be similar.
We utilize $\mathcal{L}_{mfc}$ to constrain consistency between neighbor frames, and that between each frame and the first frame, \ie,
%
\begin{equation}
\label{eq:L_cc}
\begin{split}
     \mathcal{L}_{mfc}  = \frac{1}{N-1} & \sum_{i=2}^{N}  (\left\|\mathbf{M}_{t,i} \odot ( \hat{\mathbf{I}}_{t,{i-1 \rightarrow i}} - \hat{\mathbf{I}}_{t,i})\right\|_1 
    \\ & \; \; \: + \left\|\mathbf{M}_{t,1} \odot( \hat{\mathbf{I}}_{t,{i \rightarrow 1}} - \hat{\mathbf{I}}_{t,1})\right\|_1).
\end{split}
\end{equation}
%
$\hat{\mathbf{I}}_{t,{i-1 \rightarrow i}}$ and $\hat{\mathbf{I}}_{t,{i \rightarrow 1}}$ are obtained by aligning $\hat{\mathbf{I}}_{t,{i-1}}$ to $\hat{\mathbf{I}}_{t,i}$ and aligning $\hat{\mathbf{I}}_{t,i}$ to $\hat{\mathbf{I}}_{t,1}$ with a pre-trained optical flow network~\cite{sun2018pwc}, respectively.
$\mathbf{M}_{t,i}$ and $\mathbf{M}_{t,1}$ are dynamic masks for $\hat{\mathbf{I}}_{t,i}$ and $\hat{\mathbf{I}}_{t,1}$, respectively.
Third, the blur in the lower resolution is lower level and is easier to remove~\cite{kim2022mssnet,tao2018scale}, thus the artifacts are less in models trained with down-sampled blurry video.
Taking this advantage, we impose $\mathcal{L}_{mrc}$ to assist the optimization of high-resolution models with results from low-resolution models,\ie,
%
%
\begin{equation}
\label{eq:L_sc}
   \mathcal{L}_{mrc} = ||(\mathbf{M}_{t,i})_{\downarrow}\odot((\hat{\mathbf{I}}_{t,i})_{\downarrow}- \texttt{sg}(\hat{\mathbf{I}}_{t, i}^{l}))||_1.
\end{equation}
$\hat{\mathbf{I}}_{t, i}^{l}$ is the rendered sharp image from the low-resolution model, which is pre-trained by taking the down-sampled video as supervision.
$(\cdot)_{\downarrow}$ is an image down-sampling operation. 
$\texttt{sg}$ is the stop-gradient operation.

Overall, regularization terms $\mathcal{L}_{reg}$ can be denoted as,
\begin{equation}
\label{eq:regularization}
   \mathcal{L}_{reg} = \lambda_{e}\mathcal{L}_{e} + \lambda_{mfc}\mathcal{L}_{mfc} + \lambda_{mrc}\mathcal{L}_{mrc}.
\end{equation}
$\lambda_{e}$, $\lambda_{mfc}$, and $\lambda_{mrc}$ are set to 0.1, 2, 1, respectively.
Besides, following Shape-of-Motion~\cite{wang2024shape}, we also use some other regularization terms $\mathcal{L}_{oth}$ to help reconstruct 3D motion better, and the details are in the \textit{Suppl}.

\subsection{Application to Multiple Tasks}
The blurry videos suffer from not only motion blur, but also low frame rates and scene shake generally. 
Beyond novel-view synthesis, Deblur4DGS can adjust the camera poses and timestamps to address these problems, achieving video deblurring, frame interpolation, and video stabilization.
First, when inputting camera poses of the blurry video, Deblur4DGS can render corresponding deblurring results.
Second, when feeding the interpolated camera poses and timestamps, Deblur4DGS can produce frame-interpolated results.
Third, Deblur4DGS can render a more stable video with the smoothed camera poses as inputs.

\begin{table}[t!] 
    \small
    \renewcommand\arraystretch{1}
    \begin{center}
        \vspace{-2mm}
        \scalebox{0.85}{
	\begin{tabular}{lccc}
		\toprule
	  Methods &  \tabincell{c}{PSNR$\uparrow$/SSIM$\uparrow$/LPIPS$\downarrow$ \\ $288 \times 512$ } & \tabincell{c}{PSNR$\uparrow$/SSIM$\uparrow$/LPIPS$\downarrow$ \\ $720 \times 1080$} \\
   \midrule
       DeformableGS & 15.73 / 0.623 / 0.382 & 15.55 / 0.667 / 0.421 \\
        4DGaussians & 21.98 / 0.801 / 0.197& 21.69 / 0.831 / 0.264 \\
        E-D3DGS & 23.09 / 0.830 / 0.175& 22.46 / 0.844 / 0.258 \\
        Shape-of-Motion &  26.06 / 0.910 / 0.144& 25.81 / 0.897 / 0.246 \\
        SplineGS & 26.05 / 0.901 / 0.158& 24.92 / 0.883 / 0.252 \\
        \midrule
        DyBluRF  & 26.04 / 0.916 / 0.090 & 25.71 / 0.908 / 0.159 \\
        BARD-GS & \underline{26.91} / \underline{0.923} / \underline{0.077}& \underline{26.34} / \underline{0.909} / \underline{0.139}  \\
        \midrule
        Deblur4DGS (Ours)  & \textbf{27.66} / \textbf{0.935} / \textbf{0.060}& \textbf{27.16} / \textbf{0.927} / \textbf{0.123} \\
		\bottomrule
	\end{tabular}
 }
    \end{center}
    \vspace{-3mm}
	\caption{novel view synthesis results on synthetic videos.
 }
	\label{tab:quant_novel_syn}
    \vspace{-2mm}
\end{table}

\begin{table}[t!] 
    \small
    \renewcommand\arraystretch{1}
    \begin{center}
        \vspace{-2mm}
        \scalebox{0.85}{
	\begin{tabular}{lccc}
		\toprule
	  Methods &  \tabincell{c}{CLIPIQA$\uparrow$/MUSIQ$\uparrow$ \\ Redmi } & \tabincell{c}{PSNR$\uparrow$/SSIM$\uparrow$/LPIPS$\downarrow$ \\ BARD-GS} \\
   \midrule
       DeformableGS & 0.238 / 25.903 
&  15.63 / 0.781 / 0.361  \\
        4DGaussians & 0.236 / 26.514 
&  21.32 / 0.863 / 0.221  \\
        E-D3DGS & 0.257 / 24.967 
&  22.69 / \underline{0.883} / 0.217  \\
        Shape-of-Motion &  0.277 / 24.538 
&  20.53 / 0.854 / 0.289 \\
        SplineGS & 0.252 / 32.022 
&  \textbf{23.93} / \textbf{0.899} / 0.197 \\
    \midrule
        DyBluRF  & 0.263 / 34.006
&  22.71 / 0.878 / 0.192 \\
        BARD-GS & \underline{0.288} / \underline{35.505}
&  22.69 / 0.874 / \underline{0.177}   \\
        \midrule
        Deblur4DGS (Ours)  & \textbf{0.356} / \textbf{36.756} 
&  \underline{23.10} / 0.879 / \textbf{0.161}  \\
		\bottomrule
	\end{tabular}
 }
    \end{center}
    \vspace{-3mm}
	\caption{novel view synthesis results on real-world videos.
 }
	\label{tab:quant_novel_real}
    \vspace{-4mm}
\end{table}

\begin{figure*}[t!]
    \centering
    \begin{subfigure}{0.99\textwidth}
        \ContinuedFloat
        \begin{overpic}[width=0.99\textwidth,grid=False]
        {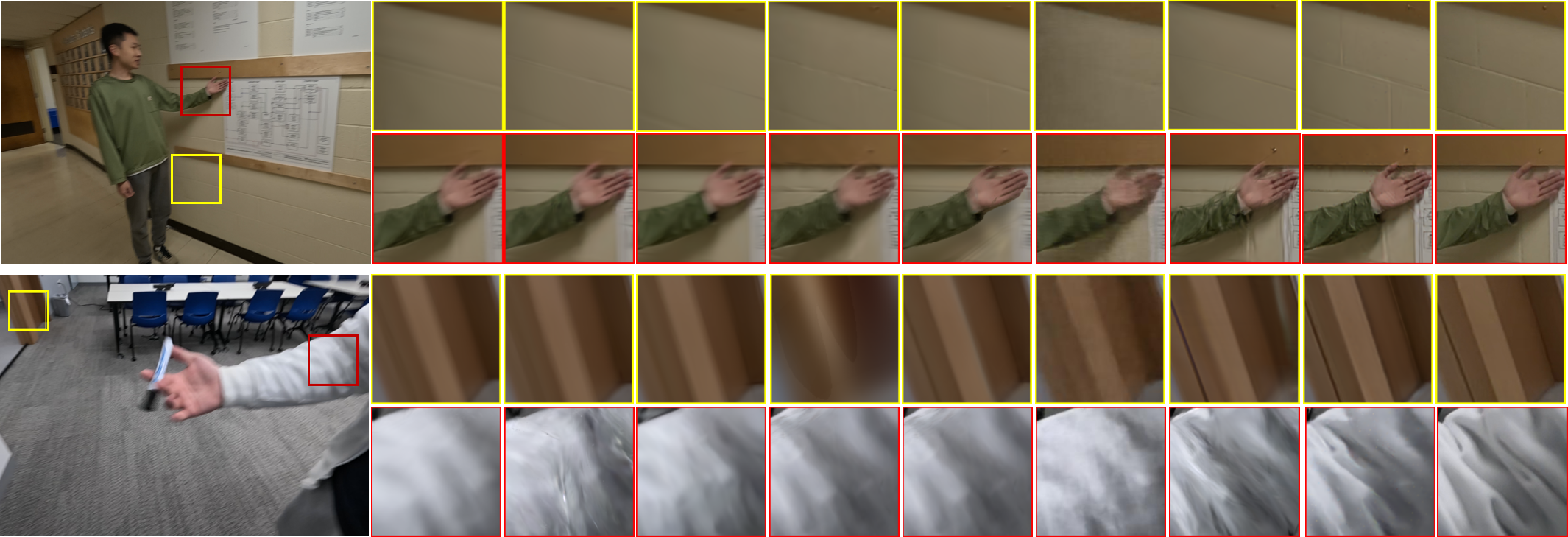}
         \put(40,-9){{\fontsize{7}{7}\selectfont Blury Frame}}
         \put(118,-9){{\fontsize{7}{7}\selectfont DeformableGS}}
         \put(163,-9){{\fontsize{7}{7}\selectfont 4DGaussians}}
         \put(204,-9){{\fontsize{7}{7}\selectfont E-D3DGS}}
         \put(240,-9){{\fontsize{7}{7}\selectfont Shape-of-Motion}}
         \put(295,-9){{\fontsize{7}{7}\selectfont SplineGS}}
         \put(335,-9){{\fontsize{7}{7}\selectfont DyBluRF}}
         \put(375,-9){{\fontsize{7}{7}\selectfont BARD-GS}}
         \put(425,-9){{\fontsize{7}{7}\selectfont Ours}}
         \put(460,-9){{\fontsize{7}{7}\selectfont Sharp GT}}
        \end{overpic}
        \vspace{1mm}
    \end{subfigure}

    \caption{Visual comparisons of novel-view synthesis on real-world videos.
    Our method produces more photo-realistic details and less visual artifacts in both static and dynamic areas, as marked with yellow and red boxes respectively.
    } 
    \label{fig:bard_vis}
    \vspace{-2mm}
\end{figure*}

\section{Experiments}
\subsection{Experimental Settings}
\noindent\textbf{Training Details.}
For stable optimization, we pre-train the camera motion predictor and static Gaussians $\mathbf{S}$ for $400$ epochs.
After that, we jointly optimize the camera motion predictor, $\mathbf{S}$, canonical dynamic Gaussians $\{\mathbf{C}_{k}\}_{k=1}^{K}$, deformable operation $\mathcal{F}$ and exposure time parameters $\{\mathbf{w}_{t}\}_{t=1}^{T}$ for $200$ epochs.
The learning rate for camera motion predictor is set to $5\times10^{-4}$ and decayed to $1\times10^{-5}$. 
The learning rate for $\{\mathbf{w}_{t}\}_{t=1}^{T}$ is set to $1\times10^{-1}$ and decayed to $1\times10^{-5}$.
The learning rate for $\mathbf{S}$, $\{\mathbf{C}_{k}\}_{k=1}^{K}$ and $\mathcal{F}$ follows Shape-of-Motion~\cite{wang2024shape}.
$N$ is set to 11.
$K$ and $H$ are set to 5 and 3 respectively.
$\mathbf{\epsilon}$ is set to $1.0$.
Experiments are conducted with PyTorch~\cite{paszke2019pytorch} on one Nvidia GeForce RTX A6000 GPU.
%


\noindent\textbf{Evaluation Configurations.}
The synthetic data contains 9 scenes with significant motion blur from Stereo Blur Dataset~\cite{zhou2019davanet}, where each scene contains blurry stereo videos and the corresponding sharp ones.
Note that DyBluRF~\cite{sun2024dyblurf}  conducts experiments on $\times 2.5$ down-sampled data.
We evaluate on both $\times 2.5$ down-sampled ones (\ie, $288 \times 512$) and the original ones (\ie, $720 \times 1080$).
For novel-view synthesis and deblurring, the rendering results may be spatially misaligned with ground truth due to the calibration error of camera parameters.
Thus, we first freeze the pre-trained 4D model and optimize camera poses by minimizing the photometric error between rendering results and ground truth, and then calculate metrics (\ie, PSNR, SSIM and LPIPS~\cite{zhang2018unreasonable}), following COLMAP-Free 3DGS~\cite{Fu_2024_CVPR}.
As there is no ground truth for frame interpolation and video stabilization, we employ recent no-reference metrics, \ie, CLIPIQA~\cite{wang2023exploring} and MUSIQ~\cite{ke2021musiq}.
Besides, we evaluate on 6 real-world blurry videos (\ie, Redmi data) captured by a Redmi K50 Ultra smartphone and 12 real-world ones (\ie, BARD-GS data) from BARD-GS~\cite{lu2025bard}, where each one contains 24 frames.
For novel-view synthesis, we employ no-reference metrics (\ie, CLIPIQA and MUSIQ) and full-reference metrics (\ie, PSNR, SSIM and LPIPS) for the two data, respectively.
For the other three tasks, we use no-reference metrics due to no ground truth.

\subsection{Comparison with State-of-the-Art Methods}
We compare with 7 state-of-the-art methods (\ie, 
DeformableGS~\cite{yang2024deformable},
4DGaussians~\cite{4dgaussians},
E-D3DGS~\cite{bae2024per}, Shape-of-Motion~\cite{wang2024shape}, SplineGS~\cite{park2025splinegs}, DyBluRF~\cite{sun2024dyblurf} and BARD-GS~\cite{lu2025bard}), where DyBluRF and BARD-GS are designed for 4D reconstruction from blurry monocular video based on NeRF and 3DGS respectively.

\begin{table}
\centering
\scalebox{0.65}{\begin{tabular}{c c c c}
\toprule
\tabincell{c}{4D \\ Methods}
  & \tabincell{c}{Deblurring \\ Methods} & \tabincell{c}{PSNR$\uparrow$ / SSIM$\uparrow$ / LPIPS$\downarrow$ \\ $288 \times 512$} & \tabincell{c}{PSNR$\uparrow$ / SSIM$\uparrow$ / LPIPS$\downarrow$ \\ $720 \times 1080$} \\
 \midrule
\multirow{4}{*}{\begin{tabular}[c]{@{}c@{}} E-D3DGS  \end{tabular}} & None & 23.09 / 0.830 / 0.175& 22.46 / 0.844 / 0.258\\
  \multicolumn{1}{c}{} & Restormer &  \textbf{23.79} / 0.850 / 0.142& 22.87 / \underline{0.864} / 0.198\\
  \multicolumn{1}{c}{} & DSTNet &  23.59 / \textbf{0.861} / \underline{0.132} & \underline{22.93} / 0.863 / \underline{0.188}\\
  \multicolumn{1}{c}{} & BSSTNet &  \underline{23.68} / \underline{0.855} / \textbf{0.128} & \textbf{23.20} / \textbf{0.877} / \textbf{0.176}\\
  \midrule
 \multirow{4}{*}{\begin{tabular}[c]{@{}c@{}} Shape-of-Motion  \end{tabular}} & None & 26.06 / 0.910 / 0.144  & 25.81 / 0.897 / 0.246 \\
  \multicolumn{1}{c}{} & Restormer & \textbf{26.80} / \textbf{0.917} / 0.085 & \textbf{26.20} / 0.911 / 0.169\\
\multicolumn{1}{c}{} & DSTNet & 26.60 / 0.915 / \underline{0.082}  & \underline{26.08} / \underline{0.914} / \underline{0.140}\\
  \multicolumn{1}{c}{} & BSSTNet& \underline{26.78} / \underline{0.916} / \textbf{0.080} & 26.06 / \textbf{0.916} / \textbf{0.125}\\
 \midrule
     Deblur4DGS   & None & \textbf{27.66} / \textbf{0.935} / \textbf{0.060}& \textbf{27.16} / \textbf{0.927} / \textbf{0.123}\\
\bottomrule
\end{tabular}}
\caption{Results that pre-process the blurry video with an image or video deblurring method before 4D reconstruction.
`None' denotes no deblurring method being used.
}
\label{tab:deblur_4d}
\vspace{-2mm}
\end{table}

\noindent\textbf{Novel-view synthesis.} 
\cref{tab:quant_novel_syn} and \cref{tab:quant_novel_real} summarize the results.
First, methods (\ie, DyBluRF and BARD-GS) that perform 4D reconstruction and motion blur modeling jointly yield overall better performance, especially in LPIPS score.
Although SplineGS gets better PSNR and SSIM scores in BARD-GS data, it produces blurry outputs.
It is consistent with the finding in BARD-GS~\cite{wang2024shape} that PSNR can sometimes yield higher values even when images appear blurrier. 
Second, benefiting the explicit 3D representation manner, BARD-GS outperforms DyBluRF, being the most competitive method.
Third, compared with BARD-GS, our Deblur4DGS performs better due to the introduction of a series of regularization terms to avoid trivial solutions and blur-aware variable canonical Gaussians to better represent dynamic objects.
Visual results in \cref{fig:real_vis} and \cref{fig:bard_vis} shows that Deblur4DGS removes blur more clearly and produces less visual artifacts in both static and dynamic areas.
Per-scene results and more visual results are in the \textit{Suppl}.
In addition, to further demonstrate the effectiveness of Deblur4DGS, we first pre-process the blurry videos with state-of-the-art image (\ie, Restormer~\cite{zamir2022restormer}) or video (DSTNet~\cite{Pan_2023_CVPR} and BSSTNet~\cite{zhang2024blur}) deblurring methods and then perform 4D reconstruction, as shown in \cref{tab:deblur_4d}.
Compared with reconstruction from blurry videos, the incorporation of deblurring models improves performance.
This is because the deblurring models remove some blur, facilitating sharp scene reconstruction.
However, as the deblurring methods cannot perceive 3D structure and maintain scene geometric consistency, the reconstruction results are still unsatisfactory.
In contrast, Deblur4DGS jointly reconstructs scene geometry and processes motion blur in 3D space, achieving better scene reconstruction results.
Visual results are in the \textit{Suppl}.

\noindent\textbf{Deblurring.}
Apart from 4D reconstruction-based methods, we compare with some state-of-the-art image and video deblurring ones.
The results are in the \textit{Suppl}.
Deblur4DGS obtains better results than 4D reconstruction-based methods and comparable ones to deblurring-specific ones.
Compared with the former, Deblur4DGS better reconstructs the scene, thus performing better.
Note that the latter ones are trained on large paired data in supervised manner while Deblur4DGS is optimized with the given blurry video in a self-supervised manner. 
Although the data prior makes them perform better, Deblur4DGS is more convenient to use.

\noindent\textbf{Frame Interpolation.}
We interpolate camera poses and timestamps to generate $\times16$ frame interpolation results.
We compare with 4D reconstruction-based methods and some video frame interpolation ones (\ie, RIFE~\cite{huang2022real}, EMAVFI~\cite{zhang2023extracting}, and VIDUE~\cite{ shang2023joint}).
The results are in the \textit{Suppl}.
VIDUE is trained with large paired data in a supervised manner for joint deblurring and fame interpolation, thus achieving better results. 
\noindent\textbf{Video Stabilization.}
We employ a Gaussian filter to smooth camera poses for video stabilization, following ~\cite{peng20243d}.
The results are in the \textit{Suppl}.
Deblur4DGS achieves pleasant scores compared with 2D video stabilization methods (\ie, MeshFlow~\cite{liu2016meshflow} and NNDVS~\cite{zhang2023minimum}) and 4D reconstruction-based ones, which benefits from better geometry reconstruction.

\begin{figure}[t!]
    \centering
    \includegraphics[width=0.9\linewidth]{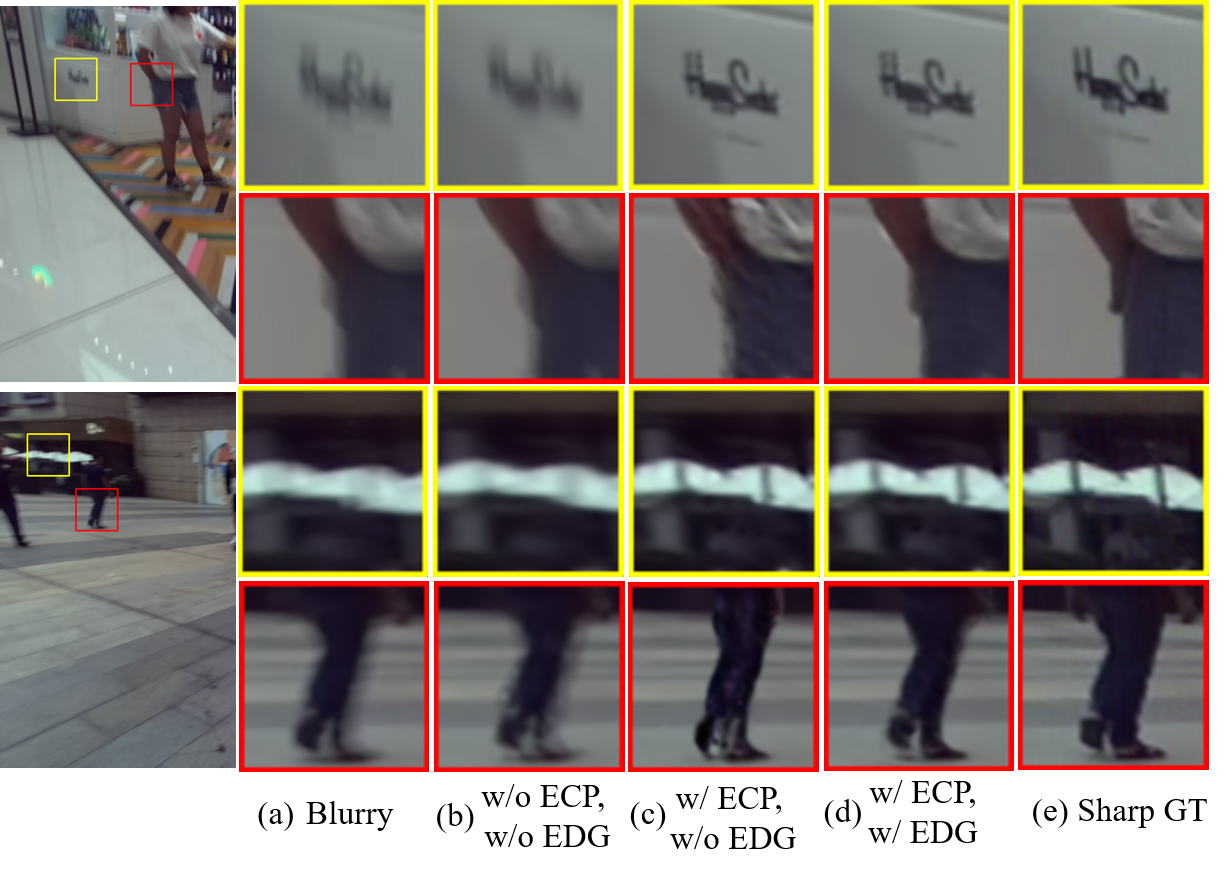}
    \vspace{-4mm}
    \caption{
    Effect of continuous camera pose (ECP) and dynamic Gaussian (EDG) estimation.
    %
    }
    \label{fig:ecp_edg}
    \vspace{-4mm}
\end{figure}

\section{Ablation Study}
We conduct experiments to validate the effectiveness of each strategy.
As our strategies in Deblur4DGS are mainly designed to process dynamic areas, we exclude the pixels of static areas to calculate the metrics in dynamic ones.

\subsection{Effect of ECP and EDG}
ECP and EDG are introduced to process camera motion blur and object motion blur, respectively.
Quantitative results and visual comparisons are shown in \cref{tab:ect_oct_dyn} and \cref{fig:ecp_edg}, respectively. 
First, without ECP and EDG, the results are almost as blurry as the input frame, as shown in \cref{fig:ecp_edg}(b).
Second, only with ECP, the static areas are sharp but may lead to visual artifacts in dynamic areas, as shown in \cref{fig:ecp_edg}(c).
It is because ECP cannot simulate the object movement.
Third, we further introduce EDG to simulate that, producing visually pleasant results in both areas, as shown in \cref{fig:ecp_edg}(d).

\subsection{Effect of Regularization Terms}
The effect of exposure regularization $\mathcal{L}_{e}$,  multi-frame consistency term $\mathcal{L}_{mfc}$ and  multi-resolution consistency term $\mathcal{L}_{mrc}$ are in \cref{tab:regularization_terms}.
Visual results are in the Sec.D of \textit{Suppl}.
Without these regularization terms, noticeable artifacts appear in dynamic regions, resulting in degraded performance.
By regularizing the object motion within the exposure time distinguished,  $\mathcal{L}_{e}$ improves performance.
Besides, $\mathcal{L}_{mfc}$ and $\mathcal{L}_{mrc}$ additionally regularize multi-frame and multi-resolution consistency respectively, helping to alleviate artifacts.
Their combinations perform best.

\subsection{Effect of BAV Canonical Gaussians.}
The results are summarized in \cref{tab:blur_awrae_canonical}.
First, selecting a single canonical Gaussians across the entire video (\ie, None) leads to poor performance, due to the challenge of modeling large object motion.
Second, selecting variable canonical Gaussians uniformly (\ie, w/o Blur-Aware) alleviates this, leading to performance gain.
We also experiment with an optical flow-based strategy~\cite{shawswings} to select canonical Gaussians, performs similar to the uniform selection.
This may be due to the inaccurate estimation of optical flow from blurry images.
Third, our blur-aware selection is better, as the canonical Gaussians from the sharper frame help blur removal.
Visual results are in the \textit{Suppl}.

\begin{table}[t!] 
\centering
\scalebox{0.75}{\begin{tabular}{c c c c}
\toprule
 ECP & EDG & \tabincell{c}{PSNR$\uparrow$/SSIM$\uparrow$/LPIPS$\downarrow$ \\ $288 \times 512$} & \tabincell{c}{PSNR$\uparrow$/SSIM$\uparrow$/LPIPS$\downarrow$ \\ $720 \times 1080$}\\
 \midrule
     $\times$  &  $\times$  &  22.10 / 0.988 / 0.018& 22.34 / 0.988 / 0.016\\
    $\checkmark$ & $\times$ & 22.30 / 0.989 / 0.016 & 22.39 / 0.988 / 0.015\\
 \midrule
    $\checkmark$  & $\checkmark$ & 22.36 / 0.989 / 0.015 & 22.63 / 0.990 / 0.014  \\
\bottomrule
\end{tabular}}
\caption{Effect about the estimation of continuous camera poses (ECP) and dynamic Gaussians (EDG).}
\label{tab:ect_oct_dyn}
\vspace{-2mm}
\end{table}

\begin{table}[t!]
\centering
\small
\scalebox{0.75}{\begin{tabular}{c c c c c}
\toprule
 $\mathcal{L}_{e}$     & $\mathcal{L}_{mfc}$      & $\mathcal{L}_{mrc}$ & \tabincell{c}{PSNR$\uparrow$/SSIM$\uparrow$/LPIPS$\downarrow$ \\ $288 \times 512$} & \tabincell{c}{PSNR$\uparrow$/SSIM$\uparrow$/LPIPS$\downarrow$ \\ $720 \times 1080$}\\
 \midrule
    $\times$&$\times$ &  $\times$ &  21.87 / 0.987 / 0.017 & 22.30 / 0.988 / 0.016\\
    $\times$&  $\checkmark$  & $\checkmark$ & 22.30 / 0.989 / 0.015 & 22.56 / 0.989 / 0.015\\
    \midrule   
    $\checkmark$ &  $\times$&  $\times$ &  22.01 / 0.988 / 0.015& 22.40 / 0.989 / 0.015\\
    $\checkmark$ & $\checkmark$ &  $\times$ & 22.16 / 0.989 / 0.016 & 22.49 / 0.989 / 0.015\\
    $\checkmark$ &  $\times$&  $\checkmark$ &   22.22 / 0.989 / 0.015&  22.54 / 0.989 / 0.014\\
  \midrule
  $\checkmark$  & $\checkmark$   & $\checkmark$&  22.36 / 0.989 / 0.015&  22.63 / 0.990 / 0.014 \\
\bottomrule
\end{tabular}}
\caption{Effetct of regularization terms (see \cref{eq:regularization}).}
\label{tab:regularization_terms}
\vspace{-2mm}
\end{table}

\begin{table}[t!]
\centering
\small
\scalebox{0.75}{\begin{tabular}{c c c c}
\toprule
 Methods & \tabincell{c}{PSNR$\uparrow$/SSIM$\uparrow$/LPIPS$\downarrow$ \\ $288 \times 512$} & \tabincell{c}{PSNR$\uparrow$/SSIM$\uparrow$/LPIPS$\downarrow$ \\ $720 \times 1080$}\\
 \midrule
  None  &  22.13 / 0.988 / 0.017& 22.30 / 0.988 / 0.016\\
  w/o Blur-Aware & 22.29 / 0.989 / 0.016 & 22.57 / 0.989 / 0.015\\
  \midrule
  Ours  &  22.36 / 0.989 / 0.015&  22.63 / 0.990 / 0.014 \\
\bottomrule
\end{tabular}}
\caption{Effect of blur-aware variable (BAV) canonical Gaussians. `None' denotes selecting a single one.
}
\label{tab:blur_awrae_canonical}
\vspace{-2mm}
\end{table}

\section{Conclusions}
In this work, we propose Deblur4DGS, a 4D Gaussian Splatting framework to reconstruct a high-quality 4D model from blurry monocular video.
In particular, with the explicit motion trajectory modeling, we propose to transform the challenging continuous dynamic representation estimation within an exposure time into the exposure time estimation, where a series of regularizations are suggested to tackle the under-constrained optimization.
Besides, a blur-aware variable canonical Gaussians is present to represent objects with large motion better.
Beyond novel-view synthesis, Deblur4DGS can improve blurry video quality from multiple perspectives, including deblurring, frame interpolation, and video stabilization.
%
Extensive results show Deblur4DGS outperforms state-of-the-art 4D reconstruction methods.

\section{Acknowledgements}
This work was supported by the National Natural Science Foundation of China under Grant No.~62371164 and the National Key RD Program of China under Grant No.~2022YFA1004100.

\bibliography{aaai2026}

\clearpage

\twocolumn[
    \begin{center}
        \huge\textbf{Supplementary Material}
    \end{center}
    \vspace{1cm}
]

\noindent{The content of the supplementary material involves:}
\vspace{1mm}
\begin{itemize}
\item Structure of camera motion predictor in Sec.A.
\vspace{1mm}
\item Other regularization terms in Sec.B.
\vspace{1mm}
\item More comparison results in Sec.C.
\vspace{1mm}
\item More ablation results in Sec.D.

\end{itemize}

\section{A Structure of Camera Motion Predictor}
\label{sec:camera_motion_predictor}
\begin{figure}[t!]
    \centering
    \includegraphics[width=0.99\linewidth]{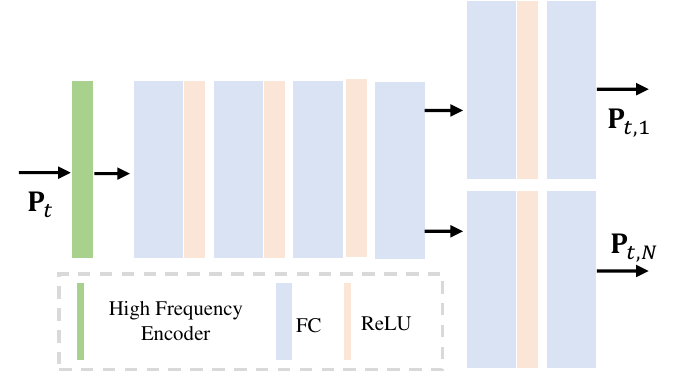}
    \caption{
          Structure of camera motion predictor. 
    }
    \label{fig:camera_motion_predictor}
\end{figure}

The structure of camera motion predictor is provided in~\cref{fig:camera_motion_predictor}.
It first embeds the camera pose $\mathbf{P}_{t}$ to a higher dimensional space using high frequency encoder~\cite{mildenhall2021nerf}.
Then, three FC blocks are stacked, each consisting of an FC layer followed by a ReLU operation.
Finally, we deploy two heads to predict the camera pose at exposure start and end (\ie, $\mathbf{P}_{t,1}$ and $\mathbf{P}_{t,N}$), respectively.

\section{B Other Regularization Terms}
\label{sec:more_regularization_terms}
Following Shape-of-Motion~\cite{wang2024shape}, we use some other regularization terms $\mathcal{L}_{oth}$ to help reconstruct 3D motion better, including  mask regularization $\mathcal{L}_{mask}$, 2D tracks regularization $\mathcal{L}_{track}$ and distance-preserving regularization $\mathcal{L}_{rigid}$.
Specifically, we render the masks within the exposure time $\{\hat{\mathbf{M}}_{t, i}\}_{i=1}^{N}$ to indicate dynamic areas.
To supervise the training, we synthesize the mask $\hat{\mathbf{M}}_{t}^{B}$ for the synthetic blurry image $\hat{\mathbf{B}}_{t}$ as,
\begin{equation}
\label{eq:syn_mask}
\hat{\mathbf{M}}_{t}^{B}(u,v) = \texttt{max}\{\hat{\mathbf{M}}_{t, 1}(u,v), \hat{\mathbf{M}}_{t, 2}(u,v),..., \hat{\mathbf{M}}_{t, N}(u,v)\}.
\end{equation}
$(u,v)$ is the pixel location.
The mask regularization $\mathcal{L}_{mask}$ can be written as,
\begin{equation}
\label{eq:mask_reg}
\mathcal{L}_{mask} = \mathcal{L}_{1}(\hat{\mathbf{M}}_{t}^{B}, \mathbf{M}_{t}),
\end{equation}
where $\mathbf{M}_{t}$ is the mask obtained by applying SAM2~\cite{ravi2024sam2} to the ground truth blurry frame.
Besides, we render the 2D tracks $\hat{\mathbf{U}}_{t \rightarrow t^\prime}$ from a pair of randomly sampled query time $t$ and target time $t^\prime$.
We supervise it by the lifted long-range 2D tracks $\mathbf{U}_{t \rightarrow t^\prime}$ that are extracted from TAPIR~\cite{doersch2023tapir}, \ie,
\begin{equation}
\label{eq:track_reg}
\mathcal{L}_{track} = \mathcal{L}_{1}(\hat{\mathbf{U}}_{t \rightarrow t^\prime}, \mathbf{U}_{t \rightarrow t^\prime}).
\end{equation}
Finally, we enforce a distance-preserving loss $\mathcal{L}_{rigid}$ between randomly sampled dynamic Gaussians and their $J$-nearest neighbors.
Let $\mathbf{x}_{t}$ and $\mathbf{x}_{t^\prime}$ denote the position of a Gaussian at time $t$ and $t^\prime$, and $\mathcal{C}_{J}(\mathbf{x}_{t})$ denote the set of $J$-nearest neighbors of $\mathbf{x}_{t}$.
$\mathcal{L}_{rigid}$ can be written as,
\begin{equation}
\label{eq:rigid_reg}
\mathcal{L}_{rigid} = ||\texttt{dist} (\hat{\mathbf{x}}_{t}, \mathcal{C}_{J}(\mathbf{x}_{t})) - \texttt{dist} (\hat{\mathbf{x}}_{t^\prime}, \mathcal{C}_{J}(\mathbf{x}_{t^\prime})) ||_{2}^{2}.
\end{equation}
$\texttt{dist}$ measures the Euclidean distance.
Overall, $\mathcal{L}_{oth}$ can be written as,
\begin{equation}
\label{eq:rigid_oth}
\mathcal{L}_{oth} = \lambda_{mask}\mathcal{L}_{mask} + \lambda_{track}\mathcal{L}_{track} + \lambda_{rigid}\mathcal{L}_{rigid}.
\end{equation}
$\lambda_{mask}$, $\lambda_{track}$ and $\lambda_{rigid}$ are set to 1, 2, and 2, respectively.

\begin{figure*}[t!]
    \centering
    \includegraphics[width=0.99\linewidth]{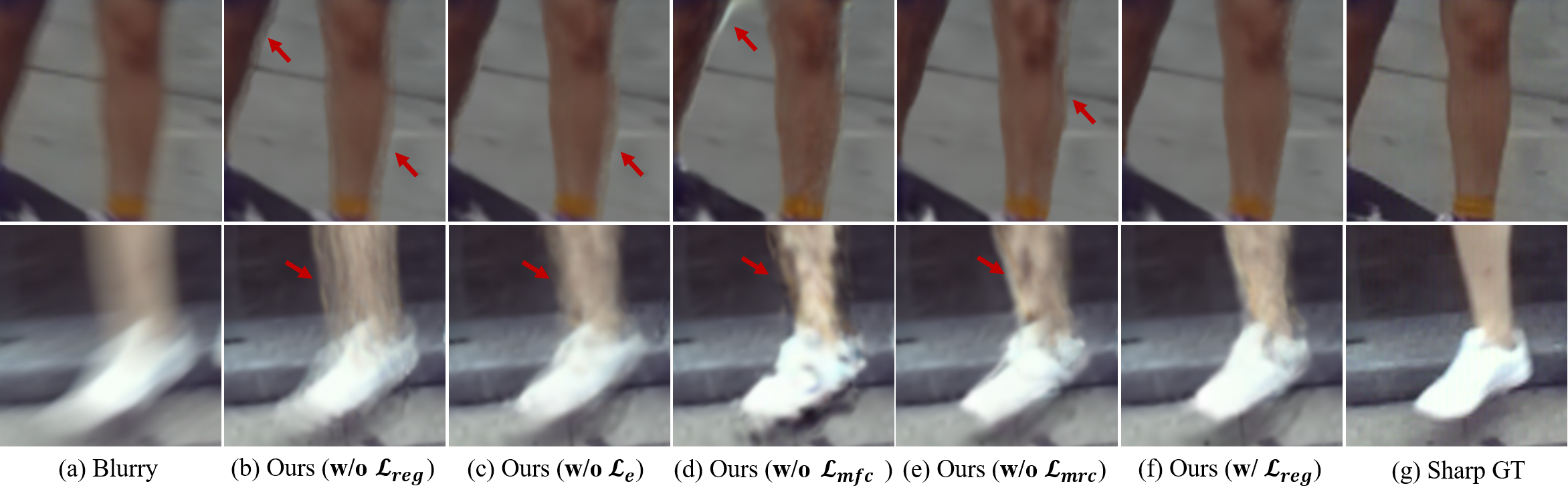}
    \vspace{-4mm}
    \caption{
    Effect of regularization terms $\mathcal{L}_{reg}$.
    $\mathcal{L}_{reg}$ includes exposure regularization $\mathcal{L}_{e}$, multi-frame
consistency regularization $\mathcal{L}_{mfc}$, and multi-resolution consistency regularization $\mathcal{L}_{mrc}$.
    }
    \label{fig:Reg_vis}
\end{figure*}

\begin{figure}[t!]
    \centering
    \includegraphics[width=0.99\linewidth]{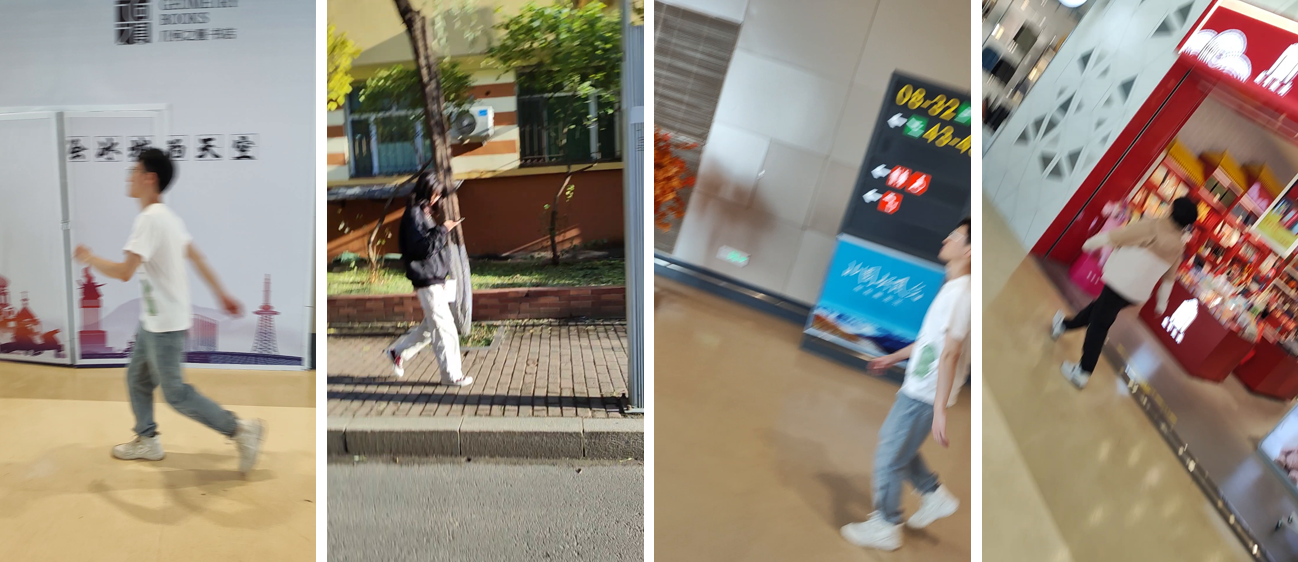}
    \caption{
          Some examples of our captured real-world blurry videos by the Redmi K50 Ultra smartphone. 
    }
    \label{fig:real_video_examples}
\end{figure}

\begin{figure}[t!]
\vspace{-1mm}
\centering
 \begin{overpic}[width=0.48\textwidth,grid=False]{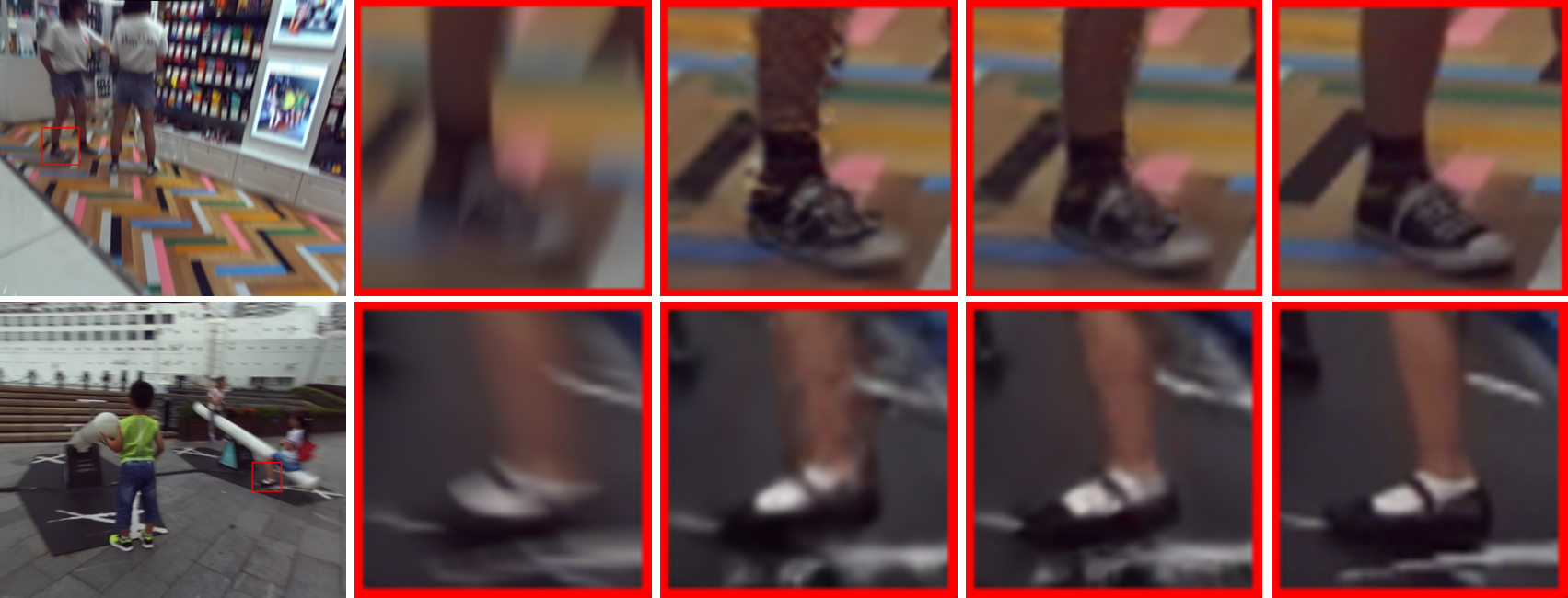}
  \put(62,-6){\tiny{Blurry Image}}
  \put(102,-6){\tiny{Ours (w/o \textit{BAV can.})}}
  \put(155,-6){\tiny{Ours (w/ \textit{BAV can.})}}
  \put(210,-6){\tiny{Sharp GT}}
\end{overpic}
\vspace{2mm}
\caption{Effect of BAV canonical Gaussians.}
\label{fig:BAV_gaussian_vis}
\end{figure}

\begin{figure}[t!]
\vspace{-1mm}
\centering
 \begin{overpic}[width=0.49\textwidth,grid=False]{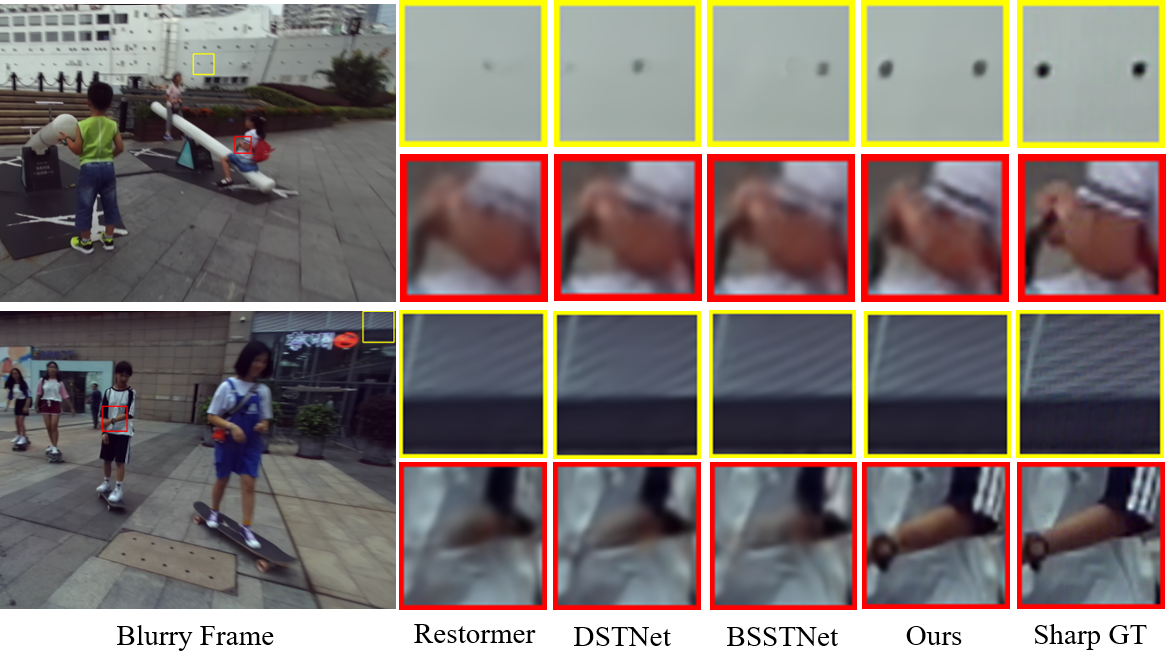}
\end{overpic}
\caption{ Visual comparisons with methods that pre-process the blurry video with an
image (\ie, Restormer~\cite{zamir2022restormer}) or video (DSTNet~\cite{Pan_2023_CVPR} and BSSTNet~\cite{zhang2024blur}) deblurring method before performing 4D reconstruction (\ie, Shape-of-Motion~\cite{wang2024shape}).}
\label{fig:deblur_som}
\end{figure}

\begin{figure}[t!]
\vspace{-1mm}
\centering
 \begin{overpic}[width=0.49\textwidth,grid=False]{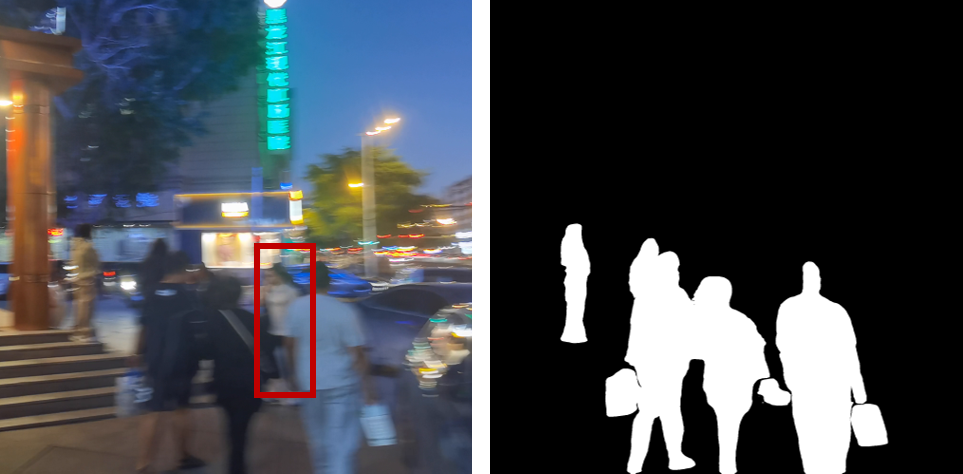}
\end{overpic}
\caption{Example of masks in extreme blurry videos. SAM 2~\cite{ravi2024sam} may fail to capture small dynamic objects under such extreme blur, as shown in the red box.}
\label{fig:extreme_blur_mask}
\end{figure}

\section{C More Comparison Results}
\cref{fig:real_video_examples} shows some examples of real-world videos captured by the Redmi K50 Ultra smartphone. 
We evaluate methods on four tasks, \ie, novel-view synthesis, deblurring, frame interpolation, and video stabilization.
Per-scene results for novel-view synthesis are summarized in \cref{tab:per_scene_288_512}, \cref{tab:per_scene_720_1280}, \cref{tab:per_scene_real} and \cref{tab:per_scene_real_bard}.\
Rendering speed comparisons are in \cref{tab:fps_compare}.
More visual comparisons for novel-view synthesis are in \cref{fig:vis_supp} and \cref{fig:real_vis}.
They show that our Deblur4DGS produces more photo-realistic details and fewer visual artifacts.
Additionally, the visual comparisons with methods that pre-process the blurry video with an
image (\ie, Restormer~\cite{zamir2022restormer}) or video (DSTNet~\cite{Pan_2023_CVPR} and BSSTNet~\cite{zhang2024blur}) deblurring method before 4D reconstruction (\ie, Shape-of-Motion~\cite{wang2024shape}) are in \cref{fig:deblur_som}, which shows that our Deblur4DGS still performs better.

\begin{figure*}[t!]
\vspace{-1mm}
\centering
 \begin{overpic}[width=0.99\textwidth,grid=False]{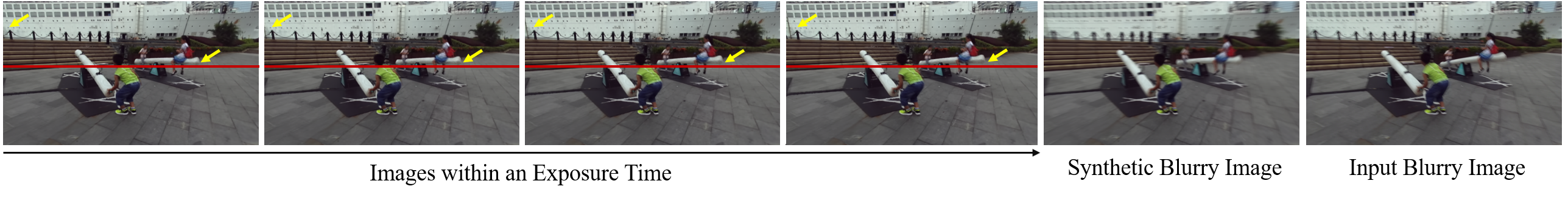}
\end{overpic}
\caption{Visual results of images within an exposure time and the synthetic blurry image. The red line is a horizontal reference line. We indicate some regions for easier observation with yellow arrows}
\label{fig:gaussian_vary}
\end{figure*}

Besides, \cref{tab:quant_deblur_synthetic}, \cref{tab:quant_fi_synthetic} and \cref{tab:quant_stab_synthetic} summarize the results for deblurring, frame interpolation, and video stabilization on synthetic videos.
\cref{tab:quant_deblur_real}, \cref{tab:quant_fi_real}, and \cref{tab:quant_stab_real} summarize those on real-world videos.
Deblur4DGS outperforms state-of-the-art 4D reconstruction methods and  has competitive capabilities in comparison with task-specific video processing models trained in a supervised manner.
Furthermore, visual results for deblurring in \cref{fig:vis_deblurring} show that compared with 4D reconstruction methods, Deblur4DGS produces sharper contents.
Moreover,  the visual results of images within an exposure time and the corresponding synthetic blurry image are in \cref{fig:gaussian_vary}.
It shows that Deblur4DGS successfully synthesize the blur by estimating camera motion and object motion trajectory.
We provide some videos at \url{https://deblur4dgs.github.io/}.

\section{D More Ablation Results}
\label{sec:more_ablation_results}
Visual results in \cref{fig:Reg_vis} validate the effectiveness of the proposed regularization terms.
Without these regularization terms, notable artifacts appear in dynamic areas.
By regularizing the object motion within the exposure time distinguished,  $\mathcal{L}_{e}$ improves performance.
Besides, $\mathcal{L}_{mfc}$ and $\mathcal{L}_{mrc}$ additionally regularize multi-frame and multi-resolution consistency respectively, helping to alleviate artifacts.
In addition, visual results in \cref{fig:BAV_gaussian_vis}  validate the effectiveness of blur-aware variable (BAV) canonical Gaussians.
Note that we extract dynamic masks using SAM 2~\cite{ravi2024sam}, which is generally robust enough to produce accurate masks even in blurry videos.
However, under extreme blur, it may fail to capture small dynamic objects, as shown in \cref{fig:extreme_blur_mask}.

\begin{table*}[t!] 
    \small
    \renewcommand\arraystretch{1}
    \begin{center}
	\caption{Per-Scene results for \textbf{novel view synthesis} on $288 \times 512$ synthetic data.}
	\label{tab:per_scene_288_512}
        \scalebox{0.62}{
	\begin{tabular}{lcccccccccc}
		\toprule
	  Methods &  \tabincell{c}{ \scriptsize{PSNR$\uparrow$/SSIM$\uparrow$/LPIPS$\downarrow$} \\\scriptsize{Skating} } & \tabincell{c}{ \scriptsize{PSNR$\uparrow$/SSIM$\uparrow$/LPIPS$\downarrow$} \\\scriptsize{Seesaw}} &
      \tabincell{c}{ \scriptsize{PSNR$\uparrow$/SSIM$\uparrow$/LPIPS$\downarrow$} \\\scriptsize{Street}} &
      \tabincell{c}{ \scriptsize{PSNR$\uparrow$/SSIM$\uparrow$/LPIPS$\downarrow$} \\\scriptsize{Basketball}} &
      \tabincell{c}{ \scriptsize{PSNR$\uparrow$/SSIM$\uparrow$/LPIPS$\downarrow$} \\\scriptsize{Children}} &
      \tabincell{c}{ \scriptsize{PSNR$\uparrow$/SSIM$\uparrow$/LPIPS$\downarrow$} \\\scriptsize{Sailor}} &
      \tabincell{c}{ \scriptsize{PSNR$\uparrow$/SSIM$\uparrow$/LPIPS$\downarrow$} \\\scriptsize{Third}} &
      \tabincell{c}{ \scriptsize{PSNR$\uparrow$/SSIM$\uparrow$/LPIPS$\downarrow$} \\\scriptsize{Man}} &
      \tabincell{c}{ \scriptsize{PSNR$\uparrow$/SSIM$\uparrow$/LPIPS$\downarrow$} \\\scriptsize{Women}}
      \\
   \midrule
DeformableGS & 16.74/0.557/0.482 & 18.27/0.695/0.237 & 12.13/0.529/0.587 & 15.34/0.645/0.306 & 15.04/0.598/0.472 & 12.89/0.491/0.539 & 16.91/0.725/0.284 & 19.48/0.747/0.273 & 14.66/0.618/0.257 \\
4DGaussians & 24.31/0.870/0.130 & 24.42/0.887/0.104 & 17.58/0.605/0.326 & 25.32/0.897/0.117 & 18.96/0.691/0.344 & 18.98/0.722/0.323 & 27.47/0.937/0.089 & 24.49/0.872/0.136 & 16.32/0.735/0.209 \\
E-D3DGS & 25.59/0.893/0.111 & 25.16/0.893/0.111 & 18.37/0.677/0.275 & 24.71/0.882/0.129 & 22.34/0.809/0.255 & 19.91/0.747/0.300 & 28.56/0.939/0.087 & 24.52/0.873/0.137 & 18.67/0.755/0.185 \\
Shape-of-Motion & \textbf{30.95}/\underline{0.959}/0.088 & 26.91/0.929/0.086 & \underline{28.56}/\underline{0.947}/0.073 & 23.31/0.865/0.180 & 23.31/0.841/0.306 & 22.07/\underline{0.894}/0.190 & 29.89/0.960/0.074 & 25.82/\underline{0.947}/0.147 & \underline{23.72}/0.851/0.151 \\
SplineGS & 28.55/0.948/0.095 & 28.48/\underline{0.953}/0.075 & 24.55/0.924/0.082 & 26.02/0.915/0.129 & 22.21/0.777/0.438 & \underline{25.31}/0.876/0.232 & 29.65/0.960/0.076 & 25.89/0.908/0.149 & \textbf{23.83}/0.848/0.129 \\
DyBluRF & 29.54/0.935/0.076 & 26.46/0.936/0.080 & 26.40/0.938/0.085 & 26.24/0.907/\textbf{0.059} & 25.32/0.903/0.111 & 25.27/0.905/\underline{0.123} & 28.57/0.937/0.070 & 24.40/0.926/0.103 & 22.23/\underline{0.860}/0.105 \\
BARD-GS & 29.78/0.940/\underline{0.060} & \underline{27.12}/0.944/\underline{0.057} & 28.25/0.939/\underline{0.067} & \underline{26.25}/\underline{0.923}/0.083 & \underline{25.67}/\underline{0.910}/\underline{0.109} & 25.21/0.901/0.111 & \underline{29.56}/\underline{0.961}/\underline{0.051} & \underline{26.87}/0.934/\underline{0.067} & 23.56/0.858/\underline{0.087} \\
Deblur4DGS (Ours) & \underline{30.84}/\textbf{0.965}/\textbf{0.047} & \textbf{28.45}/\textbf{0.954}/\textbf{0.049} & \textbf{29.01}/\textbf{0.953}/\textbf{0.048} & \textbf{27.00}/\textbf{0.932}/\underline{0.065} & \textbf{26.05}/\textbf{0.916}/\textbf{0.087} & \textbf{25.72}/\textbf{0.910}/\textbf{0.097} & \textbf{29.82}/\textbf{0.964}/\textbf{0.047} & \textbf{28.53}/\textbf{0.957}/\textbf{0.032} & 23.55/\textbf{0.865}/\textbf{0.066} \\
		\bottomrule
	\end{tabular}
 }
    \end{center}
\end{table*}

\begin{table*}[t!] 
    \small
    \renewcommand\arraystretch{1}
    \begin{center}
	\caption{Per-Scene results for \textbf{novel view synthesis} on $720 \times 1280$ synthetic data.}
	\label{tab:per_scene_720_1280}
    \scalebox{0.62}{
	\begin{tabular}{lccccccccc}
		\toprule
	  Methods &  \tabincell{c}{\scriptsize{PSNR$\uparrow$/SSIM$\uparrow$/LPIPS$\downarrow$} \\\scriptsize{Skating}} & \tabincell{c}{\scriptsize{PSNR$\uparrow$/SSIM$\uparrow$/LPIPS$\downarrow$} \\\scriptsize{Seesaw}} &
      \tabincell{c}{\scriptsize{PSNR$\uparrow$/SSIM$\uparrow$/LPIPS$\downarrow$} \\\scriptsize{Street}} &
      \tabincell{c}{\scriptsize{PSNR$\uparrow$/SSIM$\uparrow$/LPIPS$\downarrow$} \\\scriptsize{Basketball}} &
      \tabincell{c}{\scriptsize{PSNR$\uparrow$/SSIM$\uparrow$/LPIPS$\downarrow$} \\\scriptsize{Children}} &
      \tabincell{c}{\scriptsize{PSNR$\uparrow$/SSIM$\uparrow$/LPIPS$\downarrow$} \\\scriptsize{Sailor}} &
      \tabincell{c}{\scriptsize{PSNR$\uparrow$/SSIM$\uparrow$/LPIPS$\downarrow$} \\\scriptsize{Third}} &
      \tabincell{c}{\scriptsize{PSNR$\uparrow$/SSIM$\uparrow$/LPIPS$\downarrow$} \\\scriptsize{Man}} &
      \tabincell{c}{\scriptsize{PSNR$\uparrow$/SSIM$\uparrow$/LPIPS$\downarrow$} \\\scriptsize{Women}} \\
   \midrule
DeformableGS & 16.58/0.708/0.482 & 17.36/0.739/0.332 & 11.53/0.605/0.673 & 15.34/0.730/0.324 & 16.15/0.744/0.587 & 12.36/0.612/0.588 & 17.46/0.786/0.345 & 19.50/0.782/0.323 & 14.22/0.646/0.369 \\
4DGaussians & 23.75/0.880/0.203 & 23.84/0.886/0.190 & 17.99/0.740/0.344 & 20.66/0.827/0.193 & 22.32/0.830/0.398 & 18.89/0.784/0.382 & 26.99/0.930/0.168 & 23.62/0.852/0.210 & 17.16/0.752/0.286 \\
E-D3DGS & 25.09/0.897/0.186 & 24.32/0.891/0.183 & 18.04/0.764/0.353 & 23.87/0.873/0.172 & 22.27/0.833/0.394 & 19.05/0.791/0.376 & 27.37/0.927/0.181 & 23.71/0.853/0.208 & 18.43/0.771/0.276 \\
Shape-of-Motion & \underline{30.11}/\underline{0.946}/0.169 & 25.57/0.913/0.200 & 27.87/0.939/0.139 & 23.21/0.876/0.224 & 24.12/0.860/0.488 & 23.76/0.864/0.355 & \underline{28.92}/\underline{0.948}/0.156 & 24.84/0.879/0.238 & \underline{23.88}/0.853/0.248 \\
SplineGS & 28.30/0.938/0.176 & \underline{27.27}/\underline{0.936}/0.174 & 24.07/0.919/0.146 & 24.79/0.895/0.199 & 22.24/0.790/0.440 & 20.93/0.816/0.472 & 28.79/0.948/0.155 & 24.76/0.875/0.240 & 23.06/0.833/0.268 \\
DyBluRF & 29.14/0.915/0.134 & 26.56/0.930/0.128 & 26.01/0.928/0.135 & 25.84/0.902/0.147 & \underline{25.12}/\underline{0.893}/\underline{0.245} & \textbf{24.97}/\textbf{0.903}/\underline{0.220} & 27.17/0.928/\underline{0.124} & 24.30/0.919/0.151 & 22.23/0.851/0.153 \\
BARD-GS & 29.30/0.930/\underline{0.122} & 25.89/0.935/\underline{0.125} & \underline{27.98}/\underline{0.942}/\underline{0.122} & \underline{25.96}/\underline{0.912}/\underline{0.116} & 24.29/0.831/0.250 & 24.70/0.880/0.193 & 28.90/0.948/0.112 & \underline{25.84}/\underline{0.920}/\underline{0.101} & \textbf{24.20}/\textbf{0.890}/\textbf{0.113} \\
Deblur4DGS (Ours) & \textbf{30.77}/\textbf{0.956}/\textbf{0.093} & \textbf{27.52}/\textbf{0.942}/\textbf{0.107} & \textbf{28.22}/\textbf{0.944}/\textbf{0.093} & \textbf{26.93}/\textbf{0.931}/\textbf{0.109} & \textbf{25.68}/\textbf{0.905}/\textbf{0.235} & \underline{24.91}/\underline{0.897}/\textbf{0.185} & \textbf{28.96}/\textbf{0.955}/\textbf{0.104} & \textbf{28.05}/\textbf{0.948}/\textbf{0.064} & 23.50/\underline{0.865}/\underline{0.114} \\
		\bottomrule
	\end{tabular}
    }
    \end{center}
\end{table*}

\begin{table*}[t!] 
    \small
    \renewcommand\arraystretch{1}
    \begin{center}
	\caption{Per-Scene results for \textbf{novel view synthesis} on real-world videos captured by Redmi K50 Ultra.}
	\label{tab:per_scene_real}
        \scalebox{0.78}{
	\begin{tabular}{lcccccccccc}
		\toprule
	  Methods &  \tabincell{c}{CLIPIQA$\uparrow$/MUSIQ$\uparrow$ \\ Girl } & \tabincell{c}{CLIPIQA$\uparrow$/MUSIQ$\uparrow$ \\ Running} &
      \tabincell{c}{CLIPIQA$\uparrow$/MUSIQ$\uparrow$ \\Boy} &
      \tabincell{c}{CLIPIQA$\uparrow$/MUSIQ$\uparrow$ \\Walking} &
      \tabincell{c}{CLIPIQA$\uparrow$/MUSIQ$\uparrow$ \\Airport} &
      \tabincell{c}{CLIPIQA$\uparrow$/MUSIQ$\uparrow$ \\Bookshop} \\
   \midrule
       DeformableGS & 0.204/26.941& 0.190/22.321& 0.266/25.099 & 0.246/30.193 & 0.301/23.545& 0.219/27.321\\
        4DGaussians & 0.220/31.099 & 0.201/26.140 & 0.270/21.963& 0.237/27.717 & 0.267/22.642 & 0.224/29.499\\
        E-D3DGS & 0.232/30.962 & 0.219/25.331 & 0.287/23.929& 0.257/30.423& 0.282/23.707& 0.267/33.447\\
        Shape-of-Motion & \underline{0.267}/\underline{31.979} & \underline{0.234}/22.515 & 0.322/20.214& 0.241/27.609& \underline{0.344}/18.093 & 0.252/26.827\\
        SplineGS &  0.228/\textbf{48.300} & 0.207/\textbf{34.833} & 0.308/29.323 & 0.248/26.299&  0.299/25.081& 0.222/28.295 \\
        DyBluRF  & 0.169/21.756 & 0.160/29.160 & 0.303/31.051 & \underline{0.321}/25.824 & 0.218/\textbf{38.048} & 0.257/\textbf{42.090}\\
        BARD-GS &  0.245/\underline{46.379}&  0.226/30.318& \textbf{0.349}/\underline{35.593}& 0.270/\textbf{33.241}&  0.287/30.259& \textbf{0.351}/37.244\\
        Deblur4DGS (Ours)  & \textbf{0.315}/43.292 & \textbf{0.250}/\underline{31.654} & \underline{0.467}/\textbf{38.602} & \textbf{0.342}/\underline{37.978} & \textbf{0.433}/\underline{30.573}& \underline{0.327}/\underline{38.435}\\
		\bottomrule
	\end{tabular}
 }
    \end{center}
\end{table*}

\begin{table*}[t!] 
    \small
    \renewcommand\arraystretch{1.2}
    \begin{center}
	\caption{Per-Scene results for \textbf{novel view synthesis} on real-world videos from BARD-GS~\cite{lu2025bard}.}
	\label{tab:per_scene_real_bard}
    \scalebox{0.75}{
	\begin{tabular}{lcccccc}
		\toprule
		Methods & 
        \tabincell{c}{PSNR$\uparrow$/SSIM$\uparrow$/LPIPS$\downarrow$\\card} & 
        \tabincell{c}{PSNR$\uparrow$/SSIM$\uparrow$/LPIPS$\downarrow$\\cube-desk} & 
        \tabincell{c}{PSNR$\uparrow$/SSIM$\uparrow$/LPIPS$\downarrow$\\kitchen} & 
        \tabincell{c}{PSNR$\uparrow$/SSIM$\uparrow$/LPIPS$\downarrow$\\micro-lab} & 
        \tabincell{c}{PSNR$\uparrow$/SSIM$\uparrow$/LPIPS$\downarrow$\\pen-spinning} & 
        \tabincell{c}{PSNR$\uparrow$/SSIM$\uparrow$/LPIPS$\downarrow$\\poster} \\
        \midrule
        DeformableGS & 13.85/0.762/0.394 & 13.95/0.749/0.394 & 16.04/0.812/0.357 & 17.21/0.769/0.322 & 15.33/0.677/0.369 & 21.74/0.865/0.317 \\
        4DGaussians & \underline{22.39}/\underline{0.899}/0.156 & 24.69/0.896/0.207 & 21.05/0.876/0.280 & 21.34/0.852/0.249 & 18.98/0.766/0.380 & 30.34/0.945/0.163 \\
        E-D3DGS & 21.65/0.882/\underline{0.155} & \textbf{24.83}/\textbf{0.899}/0.215 & 21.52/0.888/0.285 & 22.95/\underline{0.886}/0.228 & 18.91/0.769/0.391 & \underline{30.55}/0.946/0.168 \\
        Shape-of-Motion & 18.98/0.863/0.340 & 21.26/0.884/0.308 & 21.26/0.885/0.308 & 24.80/0.882/0.234 & 15.06/0.721/0.567 & 30.50/0.950/0.154 \\
        SplineGS & \textbf{22.64}/\textbf{0.902}/\textbf{0.128} & 24.67/0.893/0.262 & \textbf{24.54}/\underline{0.913}/0.256 & \underline{25.86}/\textbf{0.897}/0.227 & 21.19/\textbf{0.826}/0.253 & \textbf{30.57}/\underline{0.951}/0.159 \\
        DyBluRF & 20.73/0.841/0.169 & 21.41/0.825/\underline{0.179} & 21.78/0.849/0.191 & \textbf{27.04}/0.863/\textbf{0.116} & \textbf{21.60}/0.652/0.345 & 30.15/0.925/\textbf{0.088} \\
        BARD-GS & 20.54/0.851/0.211 & 22.39/0.863/0.203 & 21.49/0.883/\underline{0.190} & 24.41/0.881/0.150 & \underline{21.45}/\underline{0.812}/\textbf{0.216} & 29.54/\textbf{0.954}/0.090 \\
        Deblur4DGS (Ours) & 21.56/0.882/0.189 & \underline{24.70}/\underline{0.897}/\textbf{0.155} & \underline{24.31}/\textbf{0.916}/\textbf{0.144} & 23.27/0.863/\underline{0.145} & 19.89/0.737/\underline{0.243} & 29.38/0.945/\underline{0.089} \\
        \midrule
        Methods & 
        \tabincell{c}{PSNR$\uparrow$/SSIM$\uparrow$/LPIPS$\downarrow$\\rubik-cube} & 
        \tabincell{c}{PSNR$\uparrow$/SSIM$\uparrow$/LPIPS$\downarrow$\\shark-spin} & 
        \tabincell{c}{PSNR$\uparrow$/SSIM$\uparrow$/LPIPS$\downarrow$\\tennis-ball} & 
        \tabincell{c}{PSNR$\uparrow$/SSIM$\uparrow$/LPIPS$\downarrow$\\toycar} & 
        \tabincell{c}{PSNR$\uparrow$/SSIM$\uparrow$/LPIPS$\downarrow$\\walk} & 
        \tabincell{c}{PSNR$\uparrow$/SSIM$\uparrow$/LPIPS$\downarrow$\\windmill} \\
        \midrule
        DeformableGS & 14.38/0.827/0.335 & 18.09/0.837/0.299 & 15.19/0.708/0.362 & 13.08/0.768/0.347 & 16.21/0.787/0.342 & 15.57/0.800/0.359 \\
        4DGaussians & 20.23/0.905/0.188 & 23.11/0.896/0.221 & \underline{21.35}/\underline{0.838}/0.189 & 19.10/0.887/0.155 & 25.01/0.905/0.163 & 20.62/0.792/0.257 \\
        E-D3DGS & 20.10/\underline{0.908}/0.200 & 23.07/0.897/0.235 & \textbf{22.52}/\textbf{0.852}/\underline{0.185} & \underline{20.15}/\underline{0.890}/\underline{0.145} & 24.06/0.899/0.204 & 21.50/0.894/0.204 \\
        Shape-of-Motion & 15.10/0.847/0.354 & 11.37/0.724/0.468 & 20.63/0.830/0.182 & 18.32/0.878/0.168 & 23.91/0.893/0.222 & 21.70/0.890/0.231 \\
        SplineGS & 20.78/0.907/0.196 & \underline{25.08}/\underline{0.910}/0.251 & 20.47/0.831/\textbf{0.173} & \textbf{22.35}/\textbf{0.928}/\textbf{0.095} & \underline{25.25}/\underline{0.909}/0.178 & \underline{23.67}/0.915/0.185 \\
        DyBluRF & 19.65/0.829/0.214 & 23.87/0.889/\underline{0.194} & 20.99/0.731/0.224 & 19.83/0.845/0.162 & 24.46/0.821/0.180 & 20.89/0.784/0.249 \\
        BARD-GS & \textbf{21.09}/0.900/\underline{0.169} & 23.90/0.901/0.252 & 19.16/0.745/0.196 & 19.31/0.878/0.191 & 24.51/0.895/\underline{0.138} & \textbf{24.48}/\textbf{0.927}/\textbf{0.123} \\
        Deblur4DGS (Ours) & \underline{21.05}/\textbf{0.917}/\textbf{0.156} & \textbf{25.09}/\textbf{0.915}/\textbf{0.162} & 19.96/0.777/\underline{0.179} & 18.80/0.864/0.212 & \textbf{25.60}/\textbf{0.916}/\textbf{0.112} & 23.55/\underline{0.916}/\underline{0.141} \\
		\bottomrule
	\end{tabular}
    }
    \end{center}
\end{table*}

\begin{table*}[t!]
\centering
\caption{Rendering speed comparisons on $720 \times 1080$ images.}
\label{tab:fps_compare}
\small
\scalebox{0.9}{
\begin{tabular}{lcccccccc}
\toprule
Methods & DeformableGS & 4DGaussians & E-D3DGS & Shape-of-Motion & SplineGS & DyBluRF & BARD-GS & Deblur4DGS \\
\midrule
Rendeing Speed (FPS) & 98.03 & 56.82 & 81.31 & 96.22 & 130.12 & 0.04 & 80.43 & 96.22 \\
\bottomrule
\end{tabular}}
\end{table*}

\begin{table}[t!]
\centering
\caption{\textbf{deblurring} results on synthetic datasets.}
\label{tab:quant_deblur_synthetic}
\small
\scalebox{0.85}{
\begin{tabular}{lcc}
\toprule
\textbf{Methods} & PSNR$\uparrow$/SSIM$\uparrow$/LPIPS$\downarrow$ & PSNR$\uparrow$/SSIM$\uparrow$/LPIPS$\downarrow$ \\

& 288×512 & 720×1080 \\
\midrule
Restormer & \underline{35.45} / \underline{0.984} / 0.023 & \underline{34.66} / \underline{0.976} / 0.037 \\
DSTNet & 34.79 / 0.981 / \underline{0.020} & 33.90 / 0.973 / \underline{0.034} \\
BSSTNet & \textbf{35.51} / 0.985 / \textbf{0.016} & \textbf{34.89} / 0.980 / \textbf{0.025} \\
\midrule
DeformableGS & 26.88 / 0.866 / 0.177 & 25.53 / 0.849 / 0.239 \\
4DGaussians & 29.21 / 0.916 / 0.132 & 28.41 / 0.914 / 0.201 \\
E-D3DGS & 29.79 / 0.932 / 0.122 & \underline{28.35} / 0.914 / 0.206 \\
Shape-of-Motion & 28.11 / 0.935 / 0.150 & 27.27 / 0.912 / 0.240 \\
SplineGS & 27.09 / 0.916 / 0.153 & 26.26 / 0.908 / 0.222 \\
DyBluRF & 29.44 / 0.947 / 0.081 & 28.31 / 0.916 / \underline{0.127} \\
BARD-GS & \underline{29.98} / \underline{0.949} / \underline{0.080} & 28.28 / \underline{0.919} / 0.129 \\
Deblur4DGS (Ours) & \textbf{30.36} / \textbf{0.955} / \textbf{0.078} & \textbf{29.53} / \textbf{0.929} / \textbf{0.109} \\
\bottomrule
\end{tabular}}
\end{table}

\begin{table}[t!]
\centering
\caption{\textbf{deblurring} results  on real-world datasets.}
\label{tab:quant_deblur_real}
\small
\scalebox{0.85}{
\begin{tabular}{lcc}
\toprule
\textbf{Methods} & CLIPIQA$\uparrow$/MUSIQ$\uparrow$ & CLIPIQA$\uparrow$/MUSIQ$\uparrow$ \\
 & Redmi & BARD-GS \\
\midrule
Restormer & 0.254 / 37.299 & \underline{0.278} / \underline{40.378} \\
DSTNet & \underline{0.260} / \underline{40.213} & 0.217 / 27.741 \\
BSSTNet & \textbf{0.266} / \textbf{47.639} & \textbf{0.280} / \textbf{45.798} \\
\midrule
DeformableGS & 0.249 / 26.894 & 0.325 / 31.186 \\
4DGaussians & 0.241 / 27.302 & 0.337 / 31.409 \\
E-D3DGS & 0.260 / 28.198 & 0.317 / 26.567 \\
Shape-of-Motion & \underline{0.277} / 24.846 & 0.311 / 27.872 \\
SplineGS & 0.263 / 32.044 & 0.329 / 36.453 \\
DyBluRF & 0.236 / \underline{35.240} & 0.295 / 34.792 \\
BARD-GS & 0.290 / 35.704 & \underline{0.392} / \underline{39.983} \\
Deblur4DGS (Ours) & \textbf{0.358} / \textbf{37.051} & \textbf{0.409} / \textbf{41.318} \\
\bottomrule
\end{tabular}}
\end{table}

\begin{table}[t!]
\centering
\caption{\textbf{frame interpolation} results on synthetic datasets.}
\label{tab:quant_fi_synthetic}
\small
\scalebox{0.85}{
\begin{tabular}{lcc}
\toprule
\textbf{Methods} & CLIPIQA$\uparrow$ / MUSIQ$\uparrow$ & CLIPIQA$\uparrow$ / MUSIQ$\uparrow$ \\
& 288×512 & 720×1080 \\
\midrule
RIFE               & 0.178 / \underline{41.189} & 0.156 / \underline{31.274} \\
EMAVFI             & \underline{0.179} / 39.784 & \underline{0.174} / 31.031 \\
VIDUE              & \textbf{0.263} / \textbf{61.333} & \textbf{0.186} / \textbf{49.195} \\
\midrule
DeformableGS       & 0.169 / 38.312 & 0.172 / 29.264 \\
4DGaussians        & 0.171 / 41.018 & 0.180 / 32.032 \\
E-D3DGS            & 0.173 / 39.809 & 0.184 / 31.145 \\
Shape-of-Motion    & 0.176 / 38.871 & \underline{0.202} / 31.255 \\
SplineGS           & 0.184 / 44.963 & 0.194 / 37.788 \\
DyBluRF            & 0.149 / 50.689 & 0.125 / 35.456 \\
BARD-GS            & \underline{0.198} / \underline{51.123} & 0.196 / \underline{38.256} \\
Deblur4DGS (Ours)  & \textbf{0.201} / \textbf{52.852} & \textbf{0.207} / 39.721 \\
\bottomrule
\end{tabular}}
\end{table}

\begin{table}[t!]
\centering
\caption{\textbf{frame interpolation} results on real-world datasets.}
\label{tab:quant_fi_real}
\small
\scalebox{0.85}{
\begin{tabular}{lcc}
\toprule
\textbf{Methods} & CLIPIQA$\uparrow$ / MUSIQ$\uparrow$ & CLIPIQA$\uparrow$ / MUSIQ$\uparrow$ \\
& Redmi & BARD-GS \\
\midrule
RIFE               & 0.257 / \underline{30.053} & 0.300 / \underline{30.781} \\
EMAVFI             & \underline{0.258} / 27.025 & \underline{0.318} / 30.082 \\
VIDUE              & \textbf{0.273} / \textbf{40.332} & \textbf{0.346} / \textbf{45.782} \\
\midrule
DeformableGS       & 0.242 / 25.972 & 0.320 / 30.504 \\
4DGaussians        & 0.234 / 26.923 & 0.333 / 31.010 \\
E-D3DGS            & 0.249 / 26.425 & 0.314 / 25.974 \\
Shape-of-Motion    & 0.286 / 28.078 & 0.313 / 27.809 \\
SplineGS           & 0.272 / 33.553 & 0.330 / 37.887 \\
DyBluRF            & 0.230 / 35.523 & 0.291 / 34.217 \\
BARD-GS            & \underline{0.301} / \underline{36.003} & \underline{0.390} / \underline{40.002} \\
Deblur4DGS (Ours)  & \textbf{0.360} / \textbf{37.224} & \textbf{0.416} / \textbf{41.933} \\
\bottomrule
\end{tabular}}
\end{table}

\begin{table}[t!]
\centering
\caption{\textbf{video stabilization} results on synthetic datasets.}
\label{tab:quant_stab_synthetic}
\small
\scalebox{0.85}{
\begin{tabular}{lcc}
\toprule
\textbf{Methods} & CLIPIQA$\uparrow$/MUSIQ$\uparrow$ & CLIPIQA$\uparrow$/MUSIQ$\uparrow$ \\
&288×512 & 720×1080 \\
\midrule
MeshFlow & \underline{0.154} / \underline{34.416} & \textbf{0.136} / \underline{31.105} \\
NNDVS & \textbf{0.180} / \textbf{43.003} & \underline{0.128} / \textbf{32.262} \\
\midrule
DeformableGS & 0.169 / 37.978 & 0.173 / 28.811 \\
4DGaussians & 0.151 / 41.662 & 0.177 / 32.623 \\
E-D3DGS & 0.178 / 40.773 & 0.185 / 32.362 \\
Shape-of-Motion & 0.175 / 38.971 & \underline{0.202} / 31.183 \\
SplineGS & 0.184 / 43.834 & 0.185 / 36.993 \\
DyBluRF & 0.148 / 32.091 & 0.126 / 36.115 \\
BARD-GS & 0.196 / 51.479 & 0.193 / \underline{38.472} \\
Deblur4DGS (Ours) & \textbf{0.201} / \textbf{53.060} & \textbf{0.206} / 39.786 \\
\bottomrule
\end{tabular}}
\end{table}

\begin{table}[t!]
\centering
\caption{\textbf{video stabilization} results on real-world datasets.}
\label{tab:quant_stab_real}
\small
\scalebox{0.85}{
\begin{tabular}{lcc}
\toprule
\textbf{Methods} & CLIPIQA$\uparrow$/MUSIQ$\uparrow$ & CLIPIQA$\uparrow$/MUSIQ$\uparrow$ \\
& Redmi & BARD-GS \\
\midrule
MeshFlow & \textbf{0.331} / \underline{28.650} & \textbf{0.347} / \underline{33.324} \\
NNDVS & \underline{0.254} / \textbf{28.844} & \underline{0.317} / \textbf{33.674} \\
\midrule
DeformableGS & 0.235 / 25.309 & 0.318 / 30.144 \\
4DGaussians & 0.231 / 26.472 & 0.328 / 31.487 \\
E-D3DGS & 0.253 / 27.454 & 0.317 / 26.863 \\
Shape-of-Motion & 0.269 / 24.142 & 0.311 / 27.875 \\
SplineGS & 0.259 / 31.766 & 0.328 / 36.634 \\
DyBluRF & 0.260 / 34.117 & 0.292 / 34.319 \\
BARD-GS & \underline{0.295} / \underline{35.892} & \underline{0.387} / \underline{39.275} \\
Deblur4DGS (Ours) & \textbf{0.352} / \textbf{36.351} & \textbf{0.408} / \textbf{41.317} \\
\bottomrule
\end{tabular}}
\end{table}

\clearpage

\begin{figure*}[t!]
    \centering
    \begin{subfigure}{0.99\textwidth}
        \ContinuedFloat
        \begin{overpic}[width=0.99\textwidth,grid=False]
        {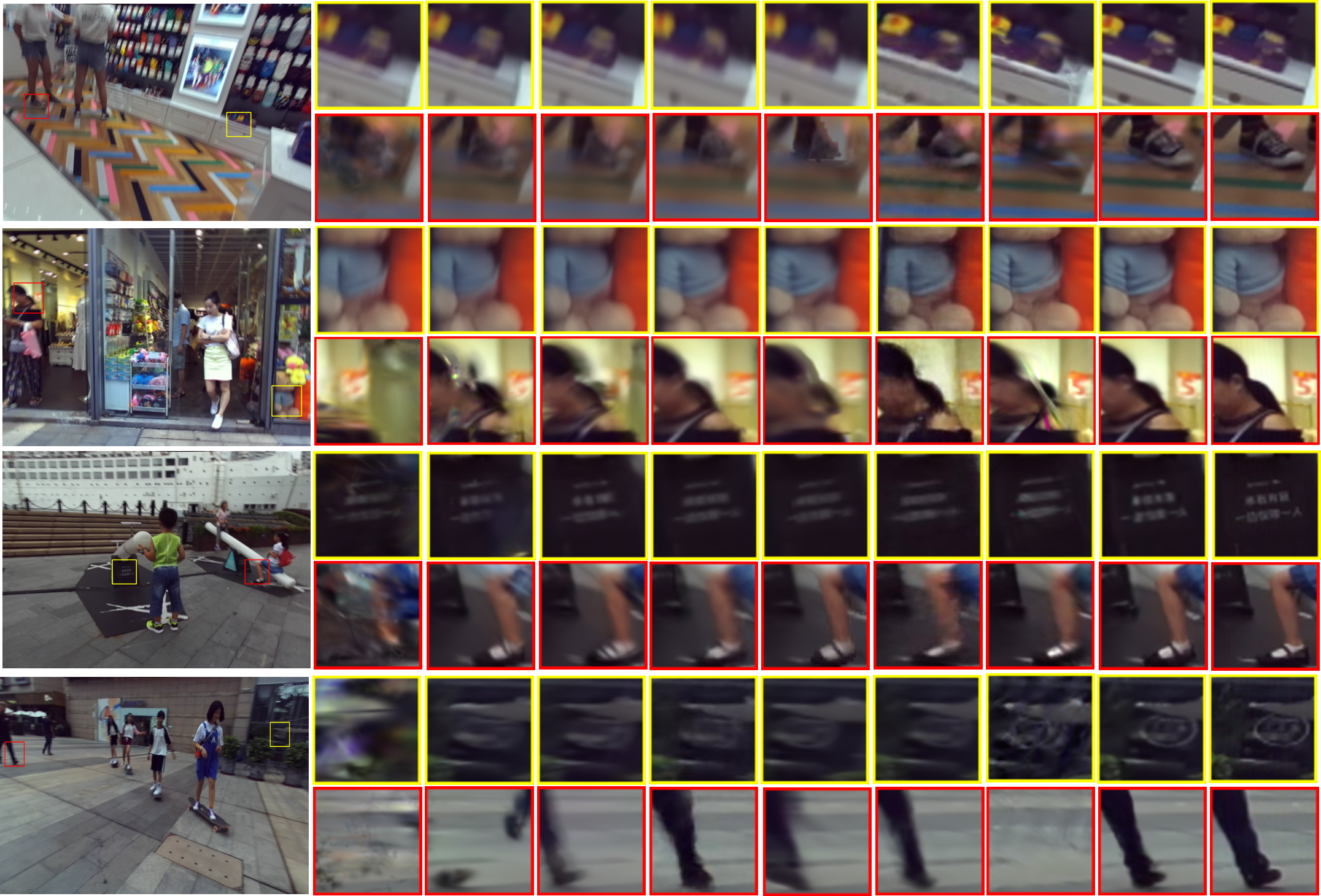}

     \put(40,-9){{\fontsize{7}{7}\selectfont Blury Frame}}
     \put(118,-9){{\fontsize{7}{7}\selectfont DeformableGS}}
     \put(163,-9){{\fontsize{7}{7}\selectfont 4DGaussians}}
     \put(204,-9){{\fontsize{7}{7}\selectfont E-D3DGS}}
     \put(240,-9){{\fontsize{7}{7}\selectfont Shape-of-Motion}}
     \put(295,-9){{\fontsize{7}{7}\selectfont SplineGS}}
     \put(335,-9){{\fontsize{7}{7}\selectfont DyBluRF}}
     \put(375,-9){{\fontsize{7}{7}\selectfont BARD-GS}}
     \put(425,-9){{\fontsize{7}{7}\selectfont Ours}}
     \put(460,-9){{\fontsize{7}{7}\selectfont Sharp GT}}
        \end{overpic}
        \vspace{2mm}
    \end{subfigure}

    \caption{Visual comparisons of novel-view synthesis on the $720 \times 1080 $ images.
    Our method produces more photo-realistic details in both static and dynamic areas, as marked with yellow and red boxes respectively.
    } 
    \label{fig:vis_supp}
\end{figure*}

\begin{figure*}[t!]
    \centering
    \begin{subfigure}{0.99\textwidth}
        \ContinuedFloat
        \begin{overpic}[width=0.99\textwidth,grid=False]
        {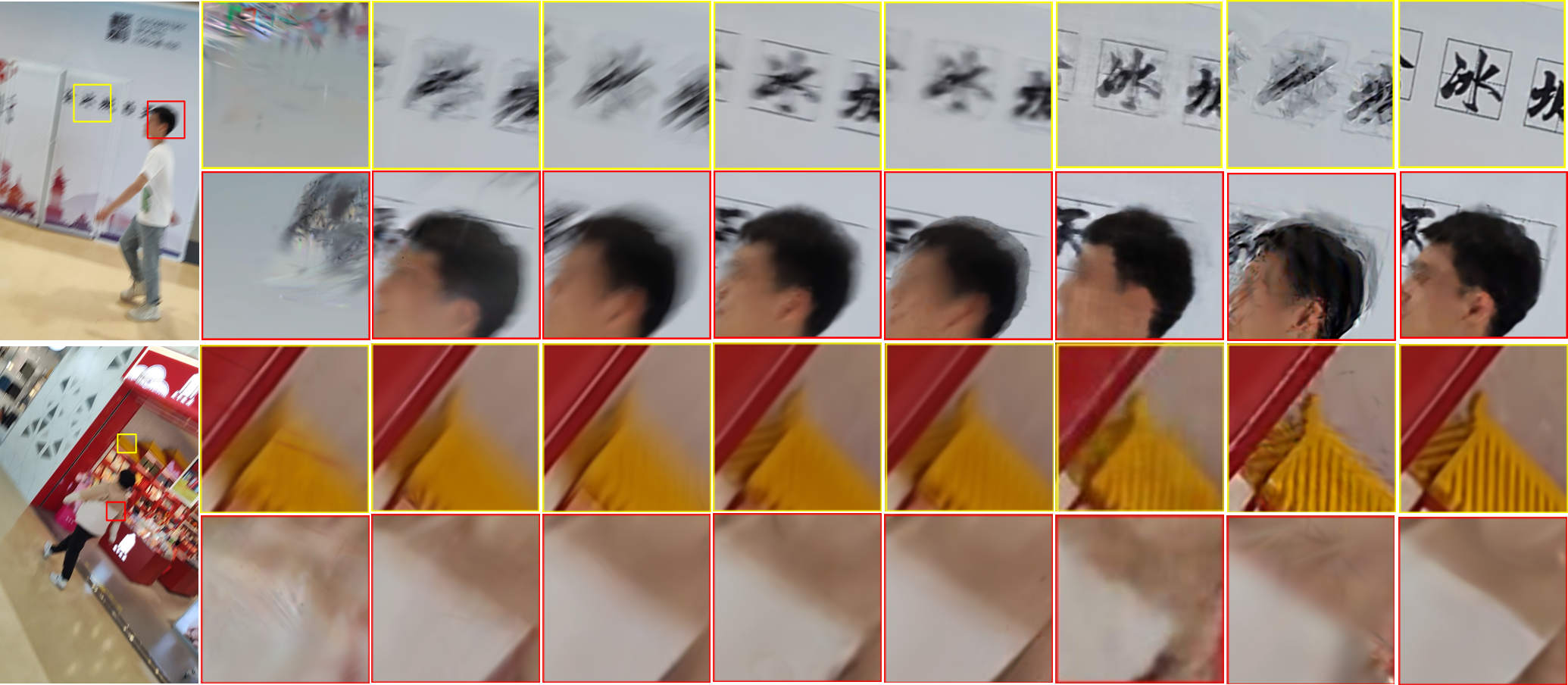}
         \put(15,-9){{\fontsize{7}{7}\selectfont Blury Frame}}
         \put(70,-9){{\fontsize{7}{7}\selectfont DeformableGS}}
         \put(125,-9){{\fontsize{7}{7}\selectfont 4DGaussians}}
         \put(185,-9){{\fontsize{7}{7}\selectfont E-D3DGS}}
         \put(228,-9){{\fontsize{7}{7}\selectfont Shape-of-Motion}}
         \put(290,-9){{\fontsize{7}{7}\selectfont SplineGS}}
         \put(348,-9){{\fontsize{7}{7}\selectfont DyBluRF}}
         \put(400,-9){{\fontsize{7}{7}\selectfont BARD-GS}}
         \put(460,-9){{\fontsize{7}{7}\selectfont Ours}}
        \end{overpic}
        \vspace{2mm}
    \end{subfigure}

    \caption{Visual comparisons of novel-view synthesis on real-world videos.
    Our method produces more photo-realistic details in both static and dynamic areas, as marked with yellow and red boxes respectively.
    } 
    \label{fig:real_vis}
\end{figure*}

\begin{figure*}[t!]
    \centering
    \begin{subfigure}{0.99\textwidth}
        \ContinuedFloat
        \begin{overpic}[width=0.99\textwidth,grid=False]
        {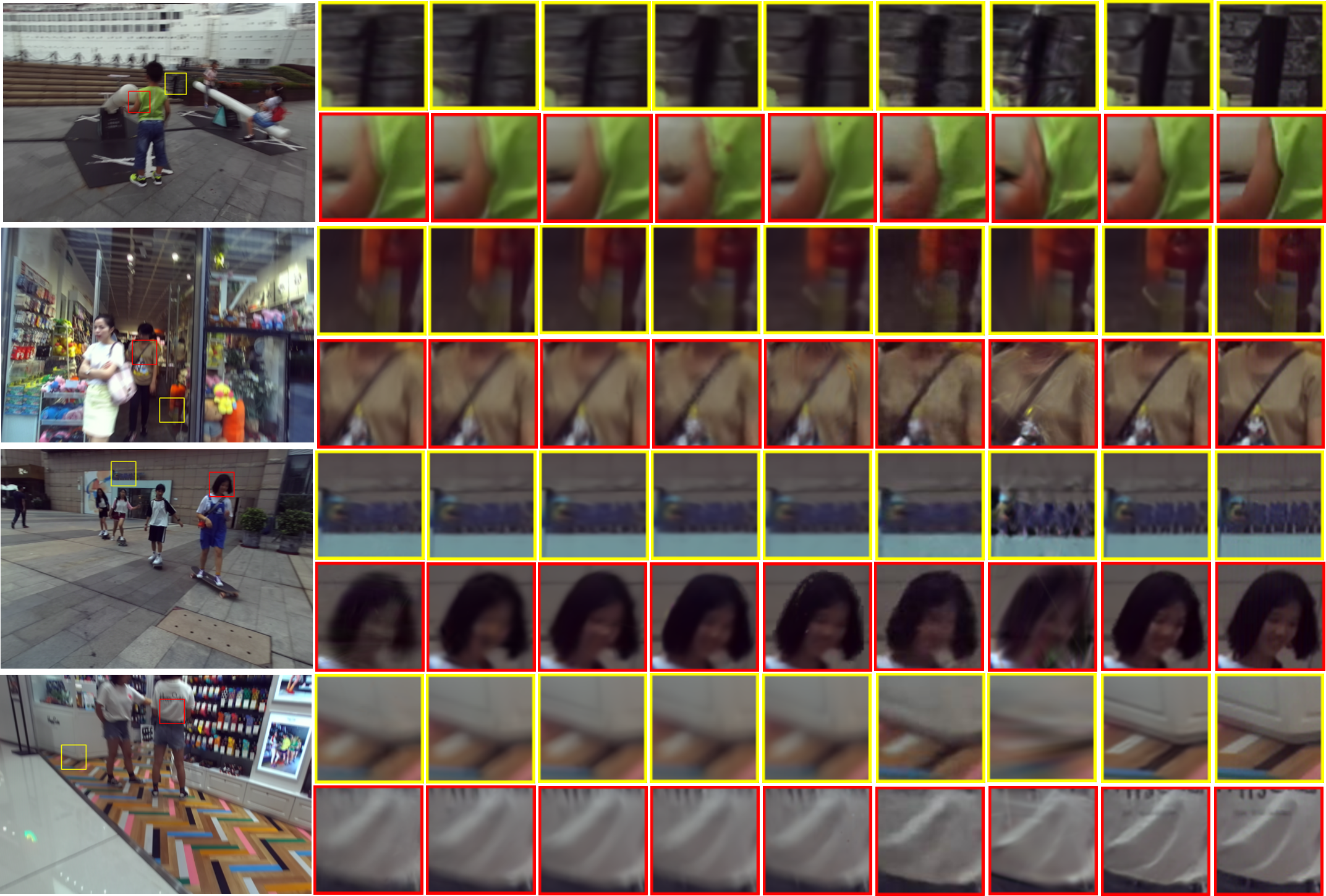}
         \put(40,-9){{\fontsize{7}{7}\selectfont Blury Frame}}
         \put(118,-9){{\fontsize{7}{7}\selectfont DeformableGS}}
         \put(163,-9){{\fontsize{7}{7}\selectfont 4DGaussians}}
         \put(204,-9){{\fontsize{7}{7}\selectfont E-D3DGS}}
         \put(240,-9){{\fontsize{7}{7}\selectfont Shape-of-Motion}}
         \put(295,-9){{\fontsize{7}{7}\selectfont SplineGS}}
         \put(335,-9){{\fontsize{7}{7}\selectfont DyBluRF}}
         \put(375,-9){{\fontsize{7}{7}\selectfont BARD-GS}}
         \put(425,-9){{\fontsize{7}{7}\selectfont Ours}}
         \put(460,-9){{\fontsize{7}{7}\selectfont Sharp GT}}
        \end{overpic}
        \vspace{2mm}
    \end{subfigure}

    \caption{Visual comparisons of deblurring on the $720 \times 1080 $ images.
    Compared with 4D reconstruction-based methods, Deblur4DGS produces sharper contents and fewer artifacts in both static and dynamic areas, as marked with yellow and red boxes respectively.
    } 
    \label{fig:vis_deblurring}
\end{figure*}

\clearpage

\end{document}